\documentclass[11pt,a4paper]{article}
\usepackage[hyperref]{acl2021}
\usepackage{times}
\usepackage{latexsym}

\usepackage{times}
\usepackage{color}
\usepackage{booktabs}
\usepackage[normalem]{ulem}
\usepackage{booktabs}
\usepackage{multirow}
\usepackage{color}
\usepackage{soul}
\usepackage{caption}
\usepackage{subcaption}
\usepackage{verbatim}
\usepackage{amssymb}

\usepackage{arabtex}

\usepackage{xurl}

\usepackage{enumitem}

\usepackage{xcolor}
\usepackage{color, soul} %用color, 和 soul 包
\usepackage{colortbl}

\usepackage{float}

\usepackage{graphicx}
\usepackage{CJKutf8}
\usepackage{multirow}
\usepackage{array}
\usepackage{tabularx}
\usepackage{makecell}
\usepackage{amsmath} 
\usepackage{amssymb}

\usepackage{amsmath} % used for boldsymbol.
\newcommand{\COMMENT}[1] {}

\usepackage{microtype}

\aclfinalcopy % Uncomment this line for the final submission
\newcolumntype{L}[1]{>{\raggedright\arraybackslash}p{#1}}
\newcolumntype{C}[1]{>{\centering\arraybackslash}p{#1}}
\newcolumntype{R}[1]{>{\raggedleft\arraybackslash}p{#1}}

\DeclareSymbolFont{extraup}{U}{zavm}{m}{n}
\DeclareMathSymbol{\varheart}{\mathalpha}{extraup}{86}
\DeclareMathSymbol{\vardiamond}{\mathalpha}{extraup}{87}

\newenvironment{abstract2}%
		 {\centerline{\large\bf Abstract}%
		  \begin{list}{}%
		     {\setlength{\rightmargin}{0.6cm}%
		      \setlength{\leftmargin}{0.6cm}}%
		   \item[]\ignorespaces%
		   }%
		 {\unskip\end{list}}

\title{ZEN 2.0: Continue Training and Adaption\\ for N-gram Enhanced Text Encoders}

\author{
    Yan Song$^{\spadesuit}$ , \hspace{0.2cm}
    Tong Zhang$^{\vardiamond}$ , \hspace{0.2cm}
    Yonggang Wang$^{\clubsuit}$ , \hspace{0.2cm}
    Kai-Fu Lee$^{\clubsuit}$\\
    $^{\spadesuit}$The Chinese University of Hong Kong (Shenzhen)\\
    $^{\vardiamond}$The Hong Kong University of Science and Technology \hspace{0.2cm}
    $^{\clubsuit}$Sinovation Ventures\\
    $^{\spadesuit}$\texttt{songyan@cuhk.edu.cn} \hspace{0.4cm}
    $^{\vardiamond}$\texttt{tongzhang@ust.hk} \\
    $^{\clubsuit}$\texttt{\{wangyonggang, kfl\}@chuangxin.com}
}

\date{}

\begin{document}

\maketitle

\begin{abstract2}

\fontsize{10pt}{12pt}\selectfont
Pre-trained text encoders have drawn sustaining attention in natural language processing (NLP) and shown their capability in obtaining promising results in different tasks.
Recent studies illustrated that external self-supervised signals (or knowledge extracted by unsupervised learning, such as n-grams) are beneficial to provide useful semantic evidence for understanding languages such as Chinese,
so as to improve the performance on various downstream tasks accordingly.
To further enhance the encoders, in this paper, we propose to pre-train n-gram-enhanced encoders with a large volume of data and advanced techniques for training.
%where careful processing is applied to data preparation as well as some effective 
%
Moreover, we try to extend the encoder to different languages as well as different domains, where it is confirmed that the same architecture is applicable to these varying circumstances and new state-of-the-art performance is observed from a long list of NLP tasks across languages and domains.
\end{abstract2}

\section{Introduction}

% intro第一段还需要在合适的地方加citation

The trend of pre-trained text encoders and decoders (language representations) has been featured in recent years with their effectiveness in performing different NLP tasks with a unified pre-training and fine-tuning paradigm \cite{devlin2019bert,dai-etal-2019-transformer,yang-2019-xlnet,wei2019nezha,Sinovation2019ZEN,baly2020arabert,joshi-etal-2020-spanbert}.
%the rapid development of NLP research and 
This paradigm, although in a ``violent manner'' with requiring huge computation cost, is nevertheless useful and provides both the academia and industry a new choice and ready-to-use resource to facilitate their NLP research and engineering.
In this context, demanding on huge data is also accompanied with the computation needs, and particularly draws attention on data quality because the pre-trained models are often learned in a self-supervised manner so that the training objectives directly depend on the data nature.
Yet, even though training data are often pre-processed and noise-filtered, some current models are restricted in learning adequate useful information from the data owing to their architecture limitations, such as applying vanilla BERT \cite{devlin2019bert} to Chinese, the important chunking information is omitted accordingly with the character-based encoder.
%即使数据很好，现有的学习框架也很难从数据中学到很intensive的知识，因此有了一些改进研究做这些事。

Therefore, many enhanced models are proposed to improve the model architecture for effective pre-training \cite{dai-etal-2019-transformer,yang-2019-xlnet,wei2019nezha,joshi-etal-2020-spanbert,liu2020k},
especially for particular language (i.e., Chinese).
Of all models, ZEN \cite{Sinovation2019ZEN} provides a flexible choice with an auxiliary encoder that learns n-gram information from the input text and uses such information to enhance the backbone character encoder.
With this design, ZEN is able to not only take the advantage of text with larger granularity, which is highly important for languages such as Chinese,
but also keep the effectiveness of the BERT architecture through its weakly supervised learning objectives.
%attaching and detaching the n-gram module.
%
Compared to other models 
% \textcolor{red}{insert some comparison analysis} 
that use different masking strategies to learn information from larger granularity without improving the model architecture \cite{cui2019pre,sun2020ernie}, ZEN explicitly encodes n-gram information and combine it into character-based encoding from the input rather than back-propagated signals from output, and thus leads to a better text representation as well as easy-to-control knowledge insertion\footnote{One can manipulate the n-gram lexicon with desired n-gram/phrases to be learned during the pre-training process.}.
In addition to the performance test in \citet{Sinovation2019ZEN}, many other studies confirmed the effectiveness of ZEN, where the state-of-the-art performance is observed on Chinese word segmentation \cite{tian2020improving}, part-of-speech (POS) tagging \cite{tian-etal-2020-joint,tian-etal-2020-joint-chinese}, parsing \cite{tian-etal-2020-constituency}, named entity recognition \cite{nie-etal-2020-improving,nie-etal-2020-named}, and conversation summarization \cite{song-etal-2020-summarizing}, when ZEN is used as the encoder.

% \cite{tian-etal-2020-joint,tian-etal-2020-joint-chinese,tian-etal-2020-constituency,tian2020improving,nie-etal-2020-improving,nie-etal-2020-named,song-etal-2020-summarizing} 

%
Although the performance of ZEN is proved on a series of Chinese NLP tasks,
it still has room for improvement on many aspects.
In doing so, there are several questions to be addressed for enhancing the current ZEN model:
(1) whether n-gram representations are still useful when continue training the model especially with its size enlarged? e.g., from base to large version;
(2) are there useful and widely applied adaptations that can be exploited by ZEN to further improve its representation ability? e.g., whole word masking;
(3) whether texts in large granularity are also informative when their representations are used to train an encoder for languages other than Chinese? e.g., for those languages that are too different from Chinese.
With such questions, we propose to update ZEN with the following improvements.
First, we propose ZEN-large, increasing the amount of its parameters to the scale of BERT-large.
Second, we refine n-gram representations with a weighting mechanism, apply whole n-gram masking and relative positional encoding during pre-training.
Third, besides Chinese, we also apply the enhanced ZEN to Arabic, which is in a different language family and greatly varies from other languages that are intensively studied in NLP, e.g., English, French, etc.

To perform such pre-training,
it is inevitable that larger models require more data and computing resources.
With the aforementioned enhancements on ZEN, we use over eight and seven billion tokens in the training corpus for Chinese and Arabic, respectively.
Especially for Arabic, to the best of our knowledge, this is the first model pre-trained exclusively for Arabic that uses such amount of data.
For the pre-training process, we use a high-performance cluster with hundreds of GPUs\footnote{All are NVidea Tesla V100 GPUs.} to perform model training and fine-tuning.
The validity and effectiveness of the enhanced ZEN are evaluated by nine widely-used tasks (with ten datasets) for Chinese and six (with ten datasets) for Arabic,
where the results confirm that a new state-of-the-art performance is achieved on these tasks.
We also analyze the factors that affect the pre-training, including training steps, weighted n-gram representations, whole n-gram masking as well as adapted character encoding, which consistently indicate that the enhancements on ZEN are effective in helping its representation ability and training efficiency.
%
%Moreover,
To facilitate the research along this line and provide useful resources to the community,
%Note that 
the enhanced ZEN is released at \url{https://github.com/sinovation/ZEN2}.
%as a new option for encoding Chinese and Arabic.

\section{ZEN 2.0}
%第一段思路：整体性的背景介绍，把ZEN拿出来先review一下，后面主要针对改进的部分和整体训练框架
%改进的部分包含以下几个方面，需要有图，所以最好有一个图可以把相关部分都展示出来

% general 地讲一些 motivation
% ZEN 1.0 是怎么做的（有什么不足）
% ZEN 2.0 有什么提升
% BERT-base: 120-130+M, ZEN 1.0 参数量 225M, ZEN 2.0 base 346M, ZEN 2.0 large 669M
% https://docs.qq.com/doc/DSlJmUkpzS1lnRUZW

% \begin{figure*}[t]
% \centering
%     \includegraphics[width=0.95\textwidth, trim=0 15 0 0 ]{figures/model_overview.png}
%     \caption{\textcolor{blue}{
%     }}
% \vskip -1em
% \label{fig: model_overview}
% \end{figure*}

Good text representations obtained from encoders often play an important role in many NLP tasks \cite{ijcai2018-607,song-etal-2017-learning,song-etal-2018-directional,devlin2019bert}.
% 
% 另外在2到2.1之间，我们要体现出我们不光针对中文，还要体现出针对阿语
% 这个得找个high-level的说法
% 
To improve character-based pre-trained encoders, ZEN 1.0 \cite{Sinovation2019ZEN} provides a framework
%
%ZEN 1.0 \cite{Sinovation2019ZEN} is a pre-trained model to improve character-based encoders (e.g., BERT \cite{devlin2019bert})
by leveraging important character-block (or text span) information with larger text granularity (i.e., n-grams) and representing such blocks with a specific encoder.\footnote{We use ``ZEN 1.0'' to refer to its original version.}
%where n-grams are used to enhance text representation.
%
In doing so,
%Specifically,
ZEN is structured with separate character and n-gram encoders.
The character encoder is a Transformer \cite{vaswani2017attention} with multiple layers following the architecture of BERT to encode input characters; 
the n-gram encoder is also a similar Transformer structure without position encoding.
When training, ZEN 1.0 follows BERT to mask several randomly selected characters in the input text.
While in both training and fine-tuning, ZEN firstly finds the n-grams in the input text according to an n-gram lexicon, in which the n-grams are text spans that are likely to contain salient contents for representing important semantic information.
Then, the model encodes these n-grams through its particular encoder and integrates their representations into the character encoder layer-wisely.
%by element-wise vector addition.
%

Based on the architecture of ZEN 1.0,
% \textcolor{blue}{
% Considering the effectiveness of ZEN framework on many NLP tasks,}
we propose an update and adaptation for this model (ZEN 2.0) from three aspects, after which the model is upgraded into the same scale of BERT-large and applied to different languages (i.e., Chinese and Arabic).
%
%In this paper, we propose ZEN 2.0
% whose architecture is illustrated in Figure \ref{fig: model_overview},
% 
First, we refine the representations of n-grams by applying weights to the n-gram representations when integrating them into the character encoder.
Second, in the training stage, we mask n-grams/words, rather than characters, in the input text of the character encoders.
Third, we utilize relative positional encoding \cite{dai-etal-2019-transformer} for the character encoder to model direction and distance information from the input text.
The details are illustrated in the following subsections.

\begin{figure}[t]
  \centering
  \includegraphics[width=0.48\textwidth, trim=0 28 0 10]{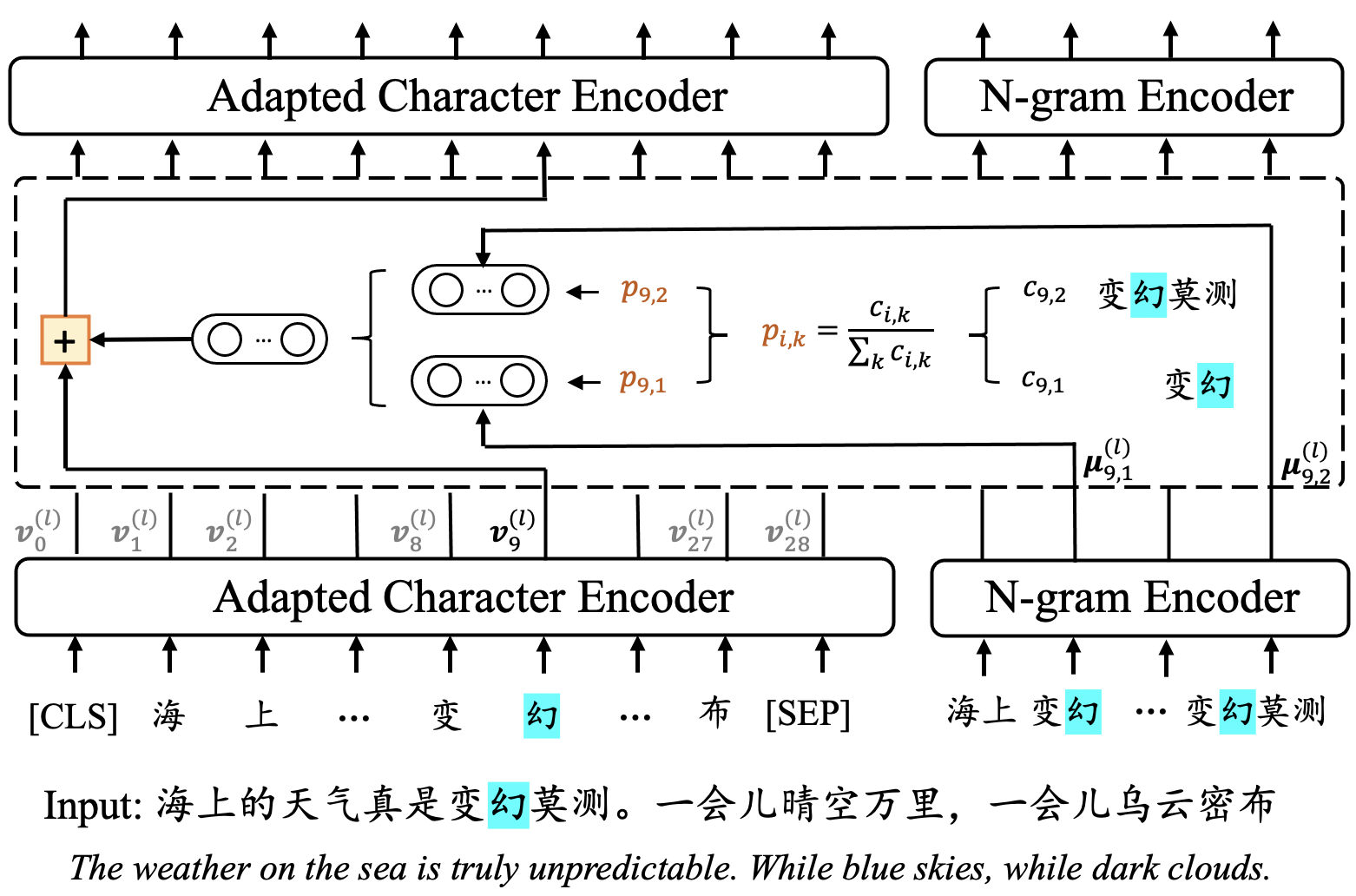}
  \caption{
  An illustration of the refined n-gram representations and their application to character encoder, where n-grams and their representations associated to the character ``\begin{CJK}{UTF8}{gkai}幻\end{CJK}'' (highlighted in blue) are weighted.
  }
  \vskip -1em
  \label{fig: ngram_representation}
\end{figure}

\begin{figure*}[t]
  \centering
  \includegraphics[width=0.98\textwidth, trim=0 25 0 20]{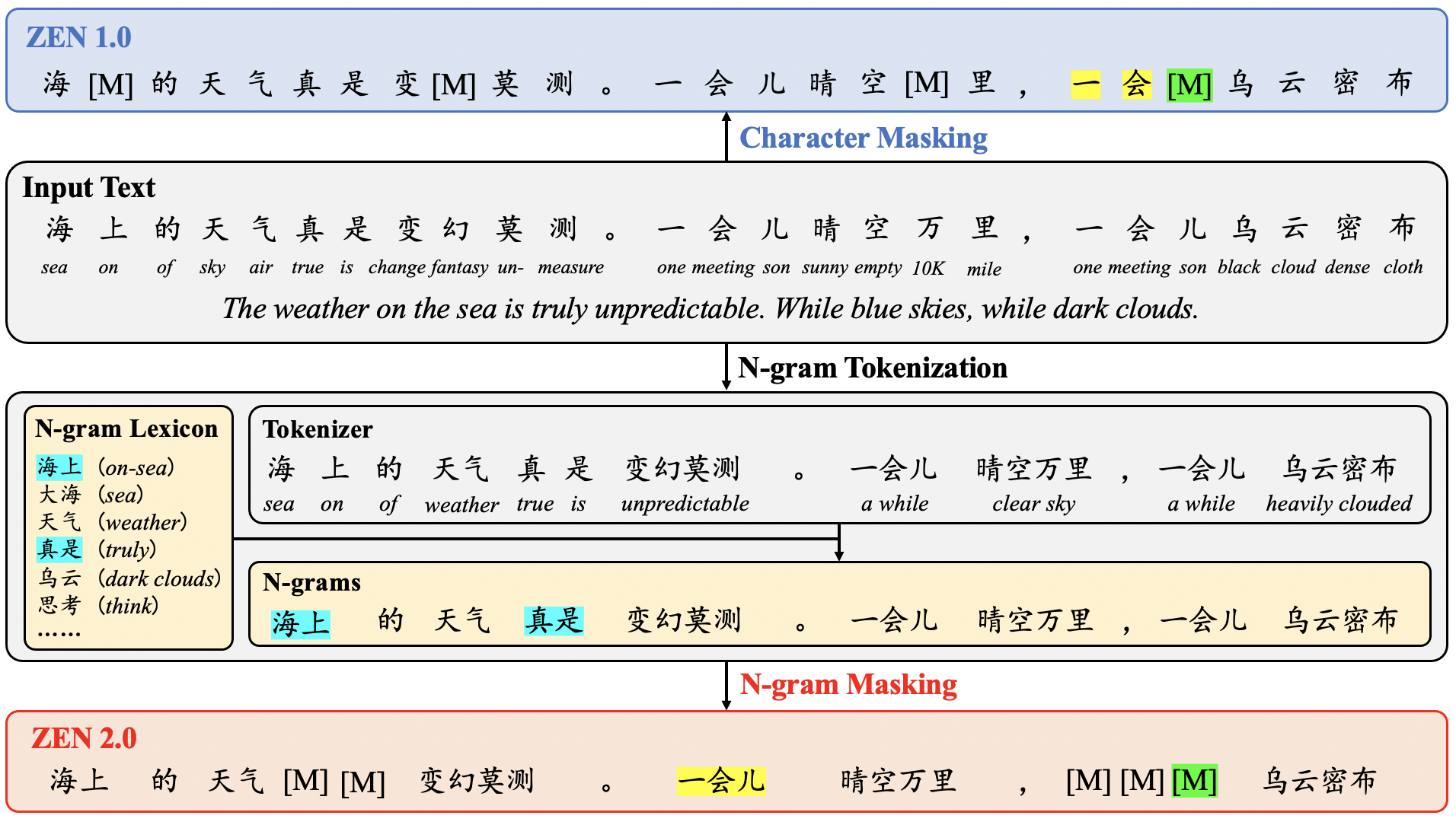}
  \caption{
  An illustration of the differences between character masking (ZEN 1.0) and n-gram masking (ZEN 2.0) with a given input text. Masked characters are represented by \texttt{[M]}.
  For ZEN 2.0,
  %In n-gram tokenization,
  adjacent character n-grams obtained from an off-the-shelf tokenizer (segmenter) are combined into a new n-gram (highlighted in blue) if that n-gram appears in the n-gram lexicon.
  %
%   In the given example,
%   compared with ZEN 1.0 that predicts the masked character ``\begin{CJK}{UTF8}{gkai}儿\end{CJK}'' (\textit{son}) highlighted in green using the information from its preceding characters ``\begin{CJK}{UTF8}{gkai}一\end{CJK}'' (\textit{one}) and ``\begin{CJK}{UTF8}{gkai}会\end{CJK}'' (\textit{meeting}) highlighted in yellow,
%   ZEN 2.0 masks larger text granularity (e.g., ``\begin{CJK}{UTF8}{gkai}一会儿\end{CJK}'' highlighted in yellow)  because ``\begin{CJK}{UTF8}{gkai}一\end{CJK}'' and ``\begin{CJK}{UTF8}{gkai}会\end{CJK}'' are parts of the valid n-gram ``\begin{CJK}{UTF8}{gkai}一会儿\end{CJK}'' so that they are treated together for masking.
% 上面是原来的
  In the given example, 
  to predict the masked character ``\begin{CJK}{UTF8}{gkai}儿\end{CJK}'' (\textit{son}) highlighted in green, ZEN 1.0 relies more on its preceding characters ``\begin{CJK}{UTF8}{gkai}一\end{CJK}'' (\textit{one}) and ``\begin{CJK}{UTF8}{gkai}会\end{CJK}'' (\textit{meeting}) highlighted in yellow (because ``\begin{CJK}{UTF8}{gkai}一会儿\end{CJK}'' (\textit{a while}) is a frequent phrase in Chinese), 
  while ZEN 2.0 is designed to learn information from large text granularity (e.g., ``\begin{CJK}{UTF8}{gkai}一会儿\end{CJK}'' highlighted in yellow in the clause ``\begin{CJK}{UTF8}{gkai}一会儿乌云密布\end{CJK}'') with all three characters masked together by whole n-gram masking.
  %by the whole n-gram masking mechanism.
  %
  %Therefore, ZEN 2.0 can learn contextual information in large text granularity and thus obtain a better text representation.
  %   
%   Characters highlighted in yellow may provide highly useful contextual information to predict the masked character highlighted in green.
  }
  \vskip -1em
  \label{fig: mask_example}
\end{figure*}

\subsection{Refined N-gram Representations}
%改进的N-gram抽取和表达，包括使用frequency对N-gram排序，以及N-gram的
%

% ZEN 1.0
% https://arxiv.org/pdf/1911.00720.pdf
% corpus: 中文 wiki
% vocab: 从大规模raw text 里面用 PMI 抽取 n-gram，然后设置一个 frequency 的阈值来 clean vocab。
% 实际使用中【抽取】 n-gram 的方法
% 随机抽取 n-gram
% 【表达】
% 公式（2），直接相加

% ZEN 2.0
% corpus: wiki, 新闻语料, 百科问答, 社区问答json版, 翻译语料,评论语料， 知乎问答
% vocab: 从大规模raw text 里面用 PMI=3 抽取 n-gram，然后设置一个 frequency=15 的阈值来 clean vocab，最后过滤n-gram长度超过8的n-gram。
% 
% 实际使用中【抽取】 n-gram 的方法
% 按照 n-gram 的 frequency 排序抽取 top-N
% 【表达】
% 公式（2），有了权重（权重是根据 frequency 来确定的，所有句子中出现的 n-gram 的 freq 做分母，该 n-gram 的 freq 做分子）

To encode n-grams, a Transformer with multi-head self-attention (MhA) are applied in ZEN 1.0.
%In ZEN 1.0, characters and n-grams are encoded through the character encoder and the n-gram encoder, respectively, where multiple layers of multi-head self-attention (MhA) are used.
%
In the process of integrating the n-gram representations into the character encoder, ZEN 1.0 enhances the representation of the $i$-th character (denoted by $\boldsymbol\upsilon_i^{(l)}$) in the $l$-th MhA layer by
\begin{equation}\label{eq: character}
\setlength\abovedisplayskip{5pt}
\setlength\belowdisplayskip{-2pt}
\boldsymbol\upsilon_i^{(l)*} =\boldsymbol\upsilon_i^{(l)} + \sum_k {\boldsymbol\mu_{i,k}^{(l)}}
\end{equation}
where $\boldsymbol\mu_{i,k}^{(l)}$ is representation of the $k$-th n-gram associated with the $i$-th character, $+$ and $\sum$ are element-wise addition operation, and
$\boldsymbol\upsilon_i^{(l)*}$ is the resulting character representation fed to the next character encoder layer.
Note that, herein, all integrated n-gram representations are treated equally.

Consider that the salience of different n-grams varies, directly summing character and n-gram representations fails to highlight the important content in particular n-grams.
Therefore, in our update of ZEN, we propose to refine n-gram representations by applying weights to original n-grams, where the process is illustrated in Figure \ref{fig: ngram_representation}.
In doing so, we adopt a simple approach by computing weights of n-grams based on their frequency of appearance in the training corpora.
Intuitively, the more frequent an n-gram is, the more likely the n-gram contains salience content when the corpus is large enough.
Then, we compute the weight $p_{i,k}$ for the $k$-th n-gram associated to the $i$-th character by
\begin{equation}\label{eq: n-gram weight}
\setlength\abovedisplayskip{5pt}
\setlength\belowdisplayskip{5pt}
p_{i,k} =\frac{c_{i,k}}{\sum_k {c_{i,k}}}
\end{equation}
where $c_{i,k}$ is the frequency of the $k$-th n-gram and $\sum_k {c_{i,k}}$ is the sum of the frequency over all n-grams associated to the $i$-th character.
Afterwards, we apply the weights to n-gram representations and obtain the results of enhanced encoding by
%ing character representation enhanced by the refined n-gram representations:
%
\begin{equation}\label{eq: refined character}
\setlength\abovedisplayskip{5pt}
\setlength\belowdisplayskip{2pt}
\boldsymbol\upsilon_i^{(l)*} = \boldsymbol\upsilon_i^{(l)} + \sum_k {p_{i,k} \cdot \boldsymbol\mu_{i,k}^{(l)}}
\end{equation}
Compared with their original form without weights, the refined n-gram representations are able to emphasize frequent n-grams and thus highlight the salient content carried by them.
% n-grams containing more important information.

\begin{figure}[t]
  \centering
  \includegraphics[width=0.46\textwidth, trim=0 25 0 10]{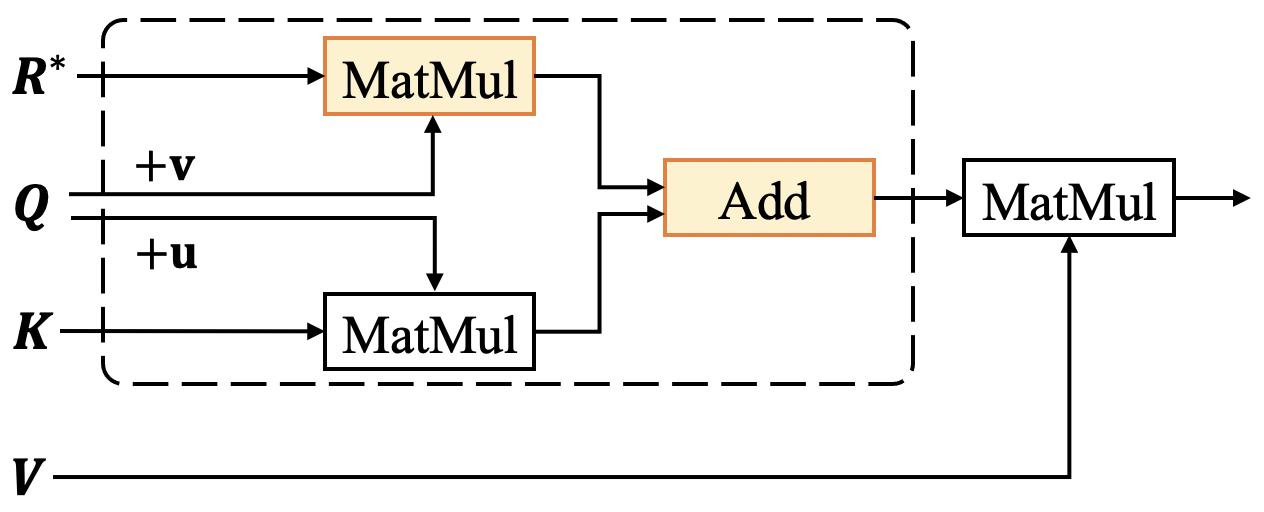}
  \caption{
  The illustration of
  the process to model the relative positional information (i.e., $\mathbf{R}^{*}$) in each head of the multi-head attention layer in the character encoder, where ``MatMul'' refers to matrix multiplication, $\mathbf{Q}$, $\mathbf{K}$, and $\mathbf{V}$ are the query, key, and value matrices, respectively, with $\mathbf{u}$ and $\mathbf{v}$ the trainable bias vectors.
  }
  \vskip -1em
  \label{fig: adapted character encoder}
\end{figure}

\subsection{Whole N-gram Masking}
%结合以往的工作，我们把whole word
% masking的思路也用到了zen2.0里面（但是在这里我们叫做whole n-gram masking）

% 原来随机 mask 字

% ZEN 2.0 
% 先用分词工具得到词（n-gram）的切分
% 随机 mask 切分得到的 word （n-gram）
% x_6 x_7 -> [MASK] [MASK]
% 在选取 [mask]、替换、还有不变的时候，是以词为单位的
% 只改变 mask 的策略，别的都是不变的

The success of whole word masking (WWM) in BERT for English indicates that it is more appropriate to mask the whole words so as to preserve (as well as predict) important semantic information carried by words rather than
%than to mask the 
sub-words/word-pieces or characters.
Motivated by WWM and the fact that word is the smallest unit that can be used in isolation with objective or practical meaning, we propose to improve ZEN 
which follows Chinese BERT to mask characters in the training stage,
by masking whole n-grams in the input text.

Because there is no natural word boundary between Chinese words in the raw text, we firstly use an off-the-shelf tokenizer to segment the input text into character n-grams and combine the adjacent ones into larger n-grams if they appear in the n-gram lexicon.
% keep those n-grams that also appear in the n-gram lexicon.
% 
Then, we randomly select some of the resulting n-grams and mask all characters in these selected n-grams, and ensure that 15\% of the characters in the input text are masked.\footnote{We follow the training procedure in original ZEN and BERT to mask 15\% of the characters in the input text.}
Figure \ref{fig: mask_example} illustrates the differences between character masking and n-gram masking in ZEN with an example input text, where the masked characters are represented by \texttt{[M]}.
Compared with character masking, n-gram masking requires to mask all characters in the same n-gram.
Afterwards, for the masked characters, we follow the conventional operation \cite{devlin2019bert,dai-etal-2019-transformer,wei2019nezha,sun2020ernie} to (1) replace 80\% of them by a special \texttt{[MASK]} token, (2) replace 10\% of them by a random token, and (3) keep 10\% of them the same.
In the training stage, ZEN with whole n-gram masking tries to predict all characters in each masked n-gram based on its context and is thus optimized accordingly by larger text units.

\subsection{Relative Positional Encoding}
%这部分解释我们怎么使用了adapted transformer在我们的字编码部分

% transformer 部分完全使用 adapted transformer 的模型架构

% 整个2.3，最好不要有跟TENET那篇paper里类似或者同样的写法和公式表达
% 2.3那里也有一个同样的问题, 还是这个adapted encoding, 这里最好不要只参考他们这一个
% 公式6和7的A字体不一样

The original
ZEN uses the same architecture (i.e., Transformer) of BERT to encode characters.
%, where multiple layers of MhA are used.
%
%
Although such encoding process is effective in most cases, it still can be improved by modeling the distance and direction information for each input when encoding them, where \citet{dai-etal-2019-transformer}
%However, the vanilla Transformer attention fails to model the distance and direction information in the encoding process \cite{dai-etal-2019-transformer}.
%\footnote{\textcolor{blue}{\citet{yan-etal-2019-tener} provide proofs in their paper}}.
% 
% As a result, \citet{dai-etal-2019-transformer} has shown that using distance- and direction-aware attention can further improve the text representation as well as model performance for many NLP tasks.
%Thus, \citet{dai-etal-2019-transformer}
proposed relative positional encoding that uses distance- and direction-aware attentions to further improve text representations, whose effectiveness is demonstrated by the decent improvement on many NLP tasks.
Owing to its effectiveness, we adopt this adaptation to ZEN for character encoding.
%\textcolor{blue}{
%Inspired by their work, we apply relative positional encoding to character encoding for another update of ZEN.}

Specifically, for the $i$-th character in the input and its context (whose indices are represented by $j$), its $d$-dimensional relative positional encoding vector is represented by $\mathbf{R}_{i-j} \in \mathbb{R}^{1 \times d}$, where, according to the positional embedding of \citet{vaswani2017attention}, the $2t$-th value in $\mathbf{R}_{i-j}$ is $sin(\frac{i-j}{10000^{2t/d}})$ and the (2$t$+1)-th value is $cos(\frac{i-j}{10000^{2t/d}})$.
For each head of self-attention, whose input $\mathbf{H}$ is a sequence of character representations, we apply three trainable matrices (i.e., $\mathbf{W}_q$, $\mathbf{W}_k$, and $\mathbf{W}_v$) to each character representation $\mathbf{H}_i$ and obtain the query vector $\mathbf{Q}_i=\mathbf{W}_q \mathbf{H}_i$, the key vector $\mathbf{K}_i=\mathbf{W}_k \mathbf{H}_i$, and the value vector $\mathbf{V}_i=\mathbf{W}_v \mathbf{H}_i$.
In addition, we also compute the relative positional representation $\mathbf{R}^*_{i-j}$ by applying a trainable matrix $\mathbf{W}_r$ to $\mathbf{R}_{i-j}$:
\begin{equation}\label{eq: character encoder, r_ij}
\setlength\abovedisplayskip{5pt}
\setlength\belowdisplayskip{5pt}
\mathbf{R}^*_{i-j}=\mathbf{W}_r \mathbf{R}_{i-j}
\end{equation}
Then, we compute the attention $\mathbf{A}_{i,j}^{\text{rel}}$ by (the process is illustrated in the dashed box in Figure \ref{fig: adapted character encoder}):%
\begin{equation}\label{eq: character encoder, a_ij}
\setlength\abovedisplayskip{5pt}
\setlength\belowdisplayskip{5pt}
\mathbf{A}_{i,j}^{\text{rel}} = (\mathbf{Q}_i + \mathbf{u}) \cdot \mathbf{K}_{j} + (\mathbf{Q}_i + \mathbf{v}) \cdot \mathbf{R}^*_{i-j}
% \mathbf{u} \cdot \mathbf{K}_{j} + \mathbf{v} \cdot \mathbf{R}^*_{i-j}
\end{equation}
where $\mathbf{u}$, $\mathbf{v}$ are two different trainable bias vectors and ``$\cdot$'' represents the inner product of two vectors.
Afterwards, we compute the output of the particular head of self-attention by
\begin{equation}\label{eq: character encoder, head output}
\setlength\abovedisplayskip{5pt}
\setlength\belowdisplayskip{5pt}
head = \text{softmax}(\mathbf{A}^{\text{rel}}) \mathbf{V}
\end{equation}
and apply it to other heads.
Finally, we follow the standard Transformer to concatenate all heads and obtain the output of MhA.

\section{Data and Training}
% 第一小节

% ZEN 2.0
% corpus: wiki, 新闻语料, 百科问答, 社区问答json版, 翻译语料,评论语料， 知乎问答
% vocab: 从大规模raw text 里面用 PMI=3 抽取 n-gram，然后设置一个 frequency=15 的阈值来 clean vocab，最后过滤n-gram长度超过8的n-gram。

\subsection{Pre-training Corpora}

\begin{table*}[t]
\begin{center}
\begin{small}
    \begin{subtable}[h]{0.52\textwidth}
    \centering
    \begin{tabular}{L{4.1cm} | C{1.2cm} | C{1.2cm} }
    \toprule
    \multicolumn{1}{c|}{\textbf{Corpora}}
    & \multicolumn{1}{c|}{\textbf{Sents \#}} 
    & \multicolumn{1}{c}{\textbf{Tokens \#}} \\
    \midrule
    Chinese Wikipedia dump & ~~~~8.8M & ~~~287.6M \\
    Chinese News Corpus & ~~52.3M & 2,307.2M \\
    Chinese Baike Corpus & ~~11.9M & ~~~335.0M \\
    Chinese Webtext Corpus & ~~27.3M & ~~~892.3M \\
    Chinese-English Parallel Corpus & ~~~~5.5M & ~~~159.4M \\
    Chinese Comments Corpus & ~~12.9M & ~~~327.0M \\
    Zhihu Corpus & 149.5M & 4,087.8M \\
    \midrule
    Total & 268.2M & 8,396.3M \\
    \bottomrule
    \end{tabular}
    \caption{Chinese Corpora}
    \end{subtable}% 
    \begin{subtable}[h]{0.46\textwidth}
    \centering
    \begin{tabular}{L{3.5cm} | C{1.2cm} | C{1.2cm} }
    \toprule
    \multicolumn{1}{c|}{\textbf{Corpora}}
    & \multicolumn{1}{c|}{\textbf{Sents \#}} 
    & \multicolumn{1}{c}{\textbf{Tokens \#}} \\
    \midrule
    Arabic Wikipedia dump & ~~~~6.3M & ~~~167.0M \\
    Arabic News Corpus & ~~~~7.2M & ~~~168.3M \\
    AraCorpus & ~~10.6M & ~~~179.8M \\
    Abu EI-Khair Corpus & ~~56.7M & 1,927.3M \\
    OSCAR & 126.6M & 4,153.0M \\
    Tashkeela & ~~~~0.5M & ~~~~~20.3M \\
    UN Parallel Corpus & ~~23.2M & ~~~675.6M \\
    \midrule
    Total & 231.1M & 7,291.3M \\
    \bottomrule
    \end{tabular}
    \caption{Arabic Corpora}
    \end{subtable}

\end{small}
\vspace{-0.4cm}
\end{center}
\caption{The statistics of Chinese (a) and Arabic (b) corpora for pre-training ZEN 2.0, where number of total sentences (Sents \#) and tokens (Tokens \#) are reported.}
\label{tab:corpus_stat}
\vskip -1em
\end{table*}

\begin{table*}[t]
\begin{center}
\begin{small}
\begin{tabular}{p{3.2cm}
                   C{1.5cm} 
                   C{0.85cm} C{0.85cm}  
                   C{0.75cm} C{0.75cm} 
                   C{0.85cm} C{0.85cm}
                   C{1.0cm} C{1.0cm} }
\toprule
\multirow{4}*{\textbf{}}
&\multicolumn{1}{c}{\textsc{\textbf {CWS}}}
&\multicolumn{2}{c}{\textsc{\textbf{POS}}}
&\multicolumn{2}{c}{\textsc{\textbf{NER}}}
&\multicolumn{2}{c}{\textsc{\textbf{DC}}}
&\multicolumn{2}{c}{\textsc{\textbf{SA}}}\\
\addlinespace[0.05cm]
\cline{2-10}
\addlinespace[0.05cm]
~
&\multicolumn{1}{c}{\textsc{\textbf {MSR-CWS}}}
&\multicolumn{2}{c}{\textsc{\textbf{CTB5}}}
&\multicolumn{2}{c}{\textsc{\textbf{MSRA-NER}}}
&\multicolumn{2}{c}{\textsc{\textbf{THUCNews}}}
&\multicolumn{2}{c}{\textsc{\textbf{ChnSentiCorp}}}\\
\addlinespace[0.05cm]
\cline{2-10}
\addlinespace[0.05cm]
~& \multicolumn{1}{c}{\small{TEST}} &
\multicolumn{1}{c}{\small{DEV}} & \multicolumn{1}{c}{\small{TEST}} &
\multicolumn{1}{c}{\small{DEV}} & \multicolumn{1}{c}{\small{TEST}} &
\multicolumn{1}{c}{\small{DEV}} & \multicolumn{1}{c}{\small{TEST}} &
\multicolumn{1}{c}{\small{DEV}} & \multicolumn{1}{c}{\small{TEST}} \\
\addlinespace[0.05cm]
\cline{2-10}
\addlinespace[0.05cm]
~& \multicolumn{1}{c}{\small{F1}} &
\multicolumn{1}{c}{\small{ACC}} & \multicolumn{1}{c}{\small{ACC}} &
\multicolumn{1}{c}{\small{F1}} & \multicolumn{1}{c}{\small{F1}} &
\multicolumn{1}{c}{\small{ACC}} & \multicolumn{1}{c}{\small{ACC}} &
\multicolumn{1}{c}{\small{ACC}} & \multicolumn{1}{c}{\small{ACC}} \\
\midrule	
ERNIE 1.0 (B)  & - & - & - & 95.00 & 93.80 & - & - & 95.20 & 95.40  \\
RoBERTa-WWM (B)  & - & - & - & - & - & 98.30 & 97.80 & 94.90 & 95.60 \\
% NEZHA (B) & - & - & - & - & - & - & - & 94.74 & 95.17 \\
NEZHA-WWM (B) & - & - & - & - & - & - & - & 94.75 & 95.84 \\
K-BERT (B) & - & - & - & 96.60 & 95.70 & - & - & 95.00 & 95.80 \\
MWA (B) & - & - & - & - & - & - & - & - & 95.52 \\
ERNIE 2.0 (B)  & - & - & - & 95.20 & 93.80 & - & - & \textbf{95.70} & 95.50 \\
MacBERT (B)  & - & - & - & - & - & 98.20 & 97.70 & 95.20 & 95.60 \\
\midrule
RoBERTa-WWM (L)  & - & - & - & - & - & 98.30 & 97.80 & 95.80 & 95.80 \\
% NEZHA (L) & - & - & - & - & - & - & - & 95.92 & 95.83 \\
NEZHA-WWM (L) & - & - & - & - & - & - & - & 95.75 & 96.00 \\
ERNIE 2.0 (L) & - & - & - & 96.30 & 95.00 & - & - & 96.10 & 95.80 \\
MacBERT (L)  & - & - & - & - & - & 98.10 & 97.90 & 95.70 & 95.90 \\
\midrule
ZEN 1.0 (B) & 98.35 & 97.43 & 96.64 & 95.95 & 95.59 & 97.66 & 97.64 & 95.66 & \textbf{96.08} \\
ZEN 1.0 (L) & 98.64 & 97.55 & 96.92 & 96.67 & 96.08 & 98.18 & 97.90 & 95.92 & 96.17 \\
ZEN 2.0 (B) & \textbf{98.42} & \textbf{97.84} & \textbf{97.00} & \textbf{95.96} & \textbf{95.54} & \textbf{97.72} & \textbf{97.64} & 94.92 & \textbf{96.08} \\
ZEN 2.0 (L) & \textbf{98.66} & \textbf{97.84} & \textbf{97.09} & \textbf{96.68} & \textbf{96.20} & \textbf{98.26} & \textbf{97.93} & \textbf{96.25} & \textbf{96.50} \\
\bottomrule
\end{tabular}

\begin{tabular}{p{3.0cm}
                   C{0.68cm} C{0.68cm} 
                   C{0.68cm} C{0.68cm} 
                   C{0.65cm} C{0.65cm} 
                   C{1.58cm} 
                   C{1.36cm} C{1.36cm} }
\toprule
\multirow{4}*{\textbf{}}
&\multicolumn{4}{c}{\textsc{\textbf{SPM}}}
&\multicolumn{2}{c}{\textsc{\textbf{NLI}}}
&\multicolumn{1}{c}{\textsc{\textbf{MRC}}}
&\multicolumn{2}{c}{\textsc{\textbf{QA}}}\\
\addlinespace[0.05cm]
\cline{2-10}
\addlinespace[0.05cm]
~
&\multicolumn{2}{c}{\textsc{\textbf{LCQMC}}}
&\multicolumn{2}{c}{\textsc{\textbf{BQ Corpus}}}
&\multicolumn{2}{c}{\textsc{\textbf{XNLI}}}
&\multicolumn{1}{c}{\textsc{\textbf{CMRC2018}}}
&\multicolumn{2}{c}{\textsc{\textbf{NLPCC-DBQA}}}\\
\addlinespace[0.05cm]
\cline{2-10}
\addlinespace[0.05cm]
~& \multicolumn{1}{c}{\small{DEV}} & \multicolumn{1}{c}{\small{TEST}} &
\multicolumn{1}{c}{\small{DEV}} & \multicolumn{1}{c}{\small{TEST}} &
\multicolumn{1}{c}{\small{DEV}} & \multicolumn{1}{c}{\small{TEST}} &
\multicolumn{1}{c}{\small{DEV}} &
\multicolumn{1}{c}{\small{DEV}} & \multicolumn{1}{c}{\small{TEST}} \\
\addlinespace[0.05cm]
\cline{2-10}
\addlinespace[0.05cm]
~& \multicolumn{1}{c}{\small{ACC}} & \multicolumn{1}{c}{\small{ACC}} &
\multicolumn{1}{c}{\small{ACC}} & \multicolumn{1}{c}{\small{ACC}} &
\multicolumn{1}{c}{\small{ACC}} & \multicolumn{1}{c}{\small{ACC}} &
\multicolumn{1}{c}{\small{EM/F1}} &
\multicolumn{1}{c}{\small{MRR/F1}} & \multicolumn{1}{c}{\small{MRR/F1}} \\
\midrule	
ERNIE 1.0 (B) & 89.70 & 87.40 & 86.10 & 84.80 & 79.90 & 78.40 & 65.10/85.10 & 95.00/82.30 & 95.10/82.70 \\
RoBERTa-WWM (B) & 89.00 & 86.40 & 86.00 & 85.00 & 80.00 & 78.80 & 67.40/87.20 & - & -  \\
% NEZHA (B) & 89.98 & 87.41 & - & - & 81.37 & 79.32 & 67.07/86.35 & - & -  \\
NEZHA-WWM (B) & 89.85 & 87.10 & - & - & \textbf{81.25} & 79.11 & 67.82/86.25 & - & - \\
K-BERT (B) & 89.20 & 87.10 & - & - & 77.20 & 77.00 & - & 94.50/-~~~~~~~ & 94.30/-~~~~~~~ \\
MWA (B) & - & 88.73 & - & - & - & 78.71 & - & - & - \\
ERNIE 2.0 (B) & \textbf{90.90} & 87.90 & \textbf{86.40} & 85.00 & 81.20 & \textbf{79.70} & 69.10/88.60 & 95.70/\textbf{84.70} & 95.70/\textbf{85.30} \\
MacBERT (B) & 89.50 & 87.00 & 86.00 & 85.20 & 80.30 & 79.30 & 68.50/87.90 & - & -  \\
\midrule
RoBERTa-WWM (L) & 90.40 & 87.00 & 86.30 & 85.80 & 82.10 & 81.20 & 68.50/88.40 & - & -  \\
% NEZHA (L) & 90.18 & 87.20 & - & - & 81.53 & 80.44 & 68.10/87.20 & - & - \\
NEZHA-WWM (L) & 90.87 & 87.94 & - & - & 82.21 & 81.17 & 67.32/86.62 & - & - \\
ERNIE 2.0 (L) & \textbf{90.90} & 87.90 & 86.50 & 85.20 & 82.60 & 81.00 & 71.50/89.90 & 95.90/85.30 & 95.80/85.80 \\
MacBERT (L) & 90.60 & 87.60 & 86.20 & 85.60 & 82.40 & 81.30 & 70.70/88.90 & - & -  \\
\midrule
ZEN 1.0 (B) & 90.20 & 87.95 & 85.75 & 85.31 & 80.48 & 79.20 & 66.51/85.72 & 93.75/81.24 & 94.14/82.88 \\
ZEN 1.0 (L) & 89.18 & 88.48 & 86.58 & 85.70 & 82.49 & 81.06 & 70.58/87.84 & 95.78/85.51 & 95.58/86.35 \\
ZEN 2.0 (B) & 89.03 & \textbf{88.71} & 86.18 & \textbf{85.42} & 79.72 & 79.30 & \textbf{70.77/87.97} & \textbf{95.90}/83.84 & \textbf{95.74}/84.43 \\
ZEN 2.0 (L) & 89.33 & \textbf{88.81} & \textbf{87.11} & \textbf{85.99} & \textbf{83.25} & \textbf{83.09} & \textbf{73.00/89.92} & \textbf{96.04/85.69} & \textbf{96.11/86.47} \\
\bottomrule
\end{tabular}
\end{small}
\vspace{-0.3cm}
\end{center}
\caption{\label{result-zh-table}
The overall performance of ZEN 2.0 base (B) and large (L) for Chinese on nine NLP tasks with the comparison against existing representative pre-trained models (with both base and large version).
%on 9 Chinese NLP tasks.
%
%``WWM'' stands for the whole word masking technique.
% NEZHA-WWM is the NEZHA model with whole word masking. RoBERTa-WWM is the RoBERTa model with whole word masking pre-trained on larger corpora. 
% For K-BERT, we report the best performance of its different version on each task.
}
\vskip -1em
\end{table*}

\begin{table*}[h]
\begin{center}
\begin{small}
\begin{tabular}{p{3.13cm}
                   C{1.05cm} C{1.05cm} 
                   C{1.05cm} C{1.05cm}
                   C{1.75cm}
                   C{1.05cm} 
                   C{1.05cm} 
                   C{1.05cm} }
\toprule
\multirow{4}*{\textbf{}}
&\multicolumn{2}{c}{\textsc{\textbf {POS}}}
&\multicolumn{3}{c}{\textsc{\textbf{NER}}}
&\multicolumn{3}{c}{\textsc{\textbf{DC}}}
\\
\addlinespace[0.05cm]
\cline{2-9}
\addlinespace[0.05cm]
~
&\multicolumn{2}{c}{\textsc{\textbf {ATB}}}
&\multicolumn{2}{c}{\textsc{\textbf{AQMAR}}}
&\multicolumn{1}{c}{\textsc{\textbf{ANERCorp}}}
&\multicolumn{1}{c}{\textsc{\textbf{AR-5}}}
&\multicolumn{1}{c}{\textsc{\textbf{AB-7}}}
&\multicolumn{1}{c}{\textsc{\textbf{KH-7}}}
\\
\addlinespace[0.05cm]
\cline{2-9}
\addlinespace[0.05cm]
~& \multicolumn{1}{c}{\small{DEV}} & \multicolumn{1}{c}{\small{TEST}} &
\multicolumn{1}{c}{\small{DEV}} & \multicolumn{1}{c}{\small{TEST}} &
\multicolumn{1}{c}{\small{TEST}} &
\multicolumn{1}{c}{\small{TEST}} &
\multicolumn{1}{c}{\small{TEST}} &
\multicolumn{1}{c}{\small{TEST}} \\
\addlinespace[0.05cm]
\cline{2-9}
\addlinespace[0.05cm]
~& \multicolumn{1}{c}{\small{ACC}} & \multicolumn{1}{c}{\small{ACC}} &
\multicolumn{1}{c}{\small{F1}} & \multicolumn{1}{c}{\small{F1}} &
\multicolumn{1}{c}{\small{F1}} & 
\multicolumn{1}{c}{\small{ACC}} &
\multicolumn{1}{c}{\small{ACC}} & 
\multicolumn{1}{c}{\small{ACC}} \\
\midrule	
Multilingual BERT (B) & 94.32 & 94.88 & 75.16 & 74.68 & 81.56 & 98.12 & 96.37 & 97.89 \\
AraBERT 0.1 (B) & 95.28 & 95.58 & 77.64 & 77.28 & 84.29 & 98.64 & \textbf{96.41} & 98.98 \\
Arabic BERT (B) & 95.44 & 95.69 & 75.24 & 75.39 & 82.34 & 98.32 & 96.00 & 98.70 \\
\midrule
Arabic BERT (L) & 95.75 & 95.92 & 78.03 & 78.49 & 85.46 & 98.75 & 96.56 & 99.05 \\
\midrule
ZEN 1.0 (B) & 96.28 & 96.24 & 77.35 & 78.21 & 84.99 & 98.49 & 96.33 & 99.10 \\
ZEN 1.0 (L) & 96.57 & 96.62 & 79.95 & 78.69 & 85.25 & 98.81 & 96.65 & \textbf{99.25} \\
ZEN 2.0 (B) & \textbf{96.43} & \textbf{96.41} & \textbf{79.91} & \textbf{78.95} & \textbf{85.34} & \textbf{98.86} & 96.33 & \textbf{99.12} \\
ZEN 2.0 (L) & \textbf{96.67} & \textbf{96.69} & \textbf{79.24} & \textbf{80.26} & \textbf{85.47} & \textbf{98.92} & \textbf{96.58} & 99.23 \\
\bottomrule
\end{tabular}

\begin{tabular}{p{3.2cm}
                   C{1.5cm} 
                   C{1.5cm} C{1.5cm} 
                   C{1.8cm} C{1.8cm}
                   C{1.8cm} }
\toprule
\multirow{4}*{\textbf{}}
&\multicolumn{1}{c}{\textsc{\textbf{SA}}}
&\multicolumn{2}{c}{\textsc{\textbf{NLI}}}
&\multicolumn{3}{c}{\textsc{\textbf{MRC}}}
\\
\addlinespace[0.05cm]
\cline{2-7}
\addlinespace[0.05cm]
~
&\multicolumn{1}{c}{\textsc{\textbf{ASTD}}}
&\multicolumn{2}{c}{\textsc{\textbf{XNLI}}}
&\multicolumn{2}{c}{\textsc{\textbf{Arabic-SQuAD}}}
&\multicolumn{1}{c}{\textsc{\textbf{ARCD}}}
\\
\addlinespace[0.05cm]
\cline{2-7}
\addlinespace[0.05cm]
~& \multicolumn{1}{c}{\small{TEST}} &
\multicolumn{1}{c}{\small{DEV}} & \multicolumn{1}{c}{\small{TEST}} &
\multicolumn{1}{c}{\small{DEV}} & \multicolumn{1}{c}{\small{TEST}} &
\multicolumn{1}{c}{\small{TEST}} \\
\addlinespace[0.05cm]
\cline{2-7}
\addlinespace[0.05cm]
~& \multicolumn{1}{c}{\small{ACC}} &
\multicolumn{1}{c}{\small{ACC}} & \multicolumn{1}{c}{\small{ACC}} & 
\multicolumn{1}{c}{\small{EM/F1}} & \multicolumn{1}{c}{\small{EM/F1}} &
\multicolumn{1}{c}{\small{EM/F1}} \\
\midrule	
Multilingual BERT (B) & 68.93 & 70.72 & 71.39 & 40.42/56.78 & 40.44/56.81 & 26.49/54.50  \\
AraBERT 0.1 (B) & 72.67 & 74.43 & 74.99 & 40.12/56.91 & 40.05/56.78 & 22.93/52.99 \\
Arabic BERT (B) & 70.57 & 73.86 & 73.49 & 35.34/51.78 & 35.00/51.11 & 18.09/44.45 \\
\midrule
Arabic BERT (L) & 71.22 & 76.02 & 76.29 & 41.74/58.90 & 40.93/56.40 & 28.06/57.67 \\
\midrule
ZEN 1.0 (B) & 72.72 & 78.71 & 78.42 & 42.60/59.20 & 42.65/58.31 & 28.63/58.09 \\
ZEN 1.0 (L) & 74.27 & 82.65 & 82.34 & \textbf{47.77/64.35} & 45.89/62.33 & 36.01/67.58 \\
ZEN 2.0 (B) & \textbf{73.17} & \textbf{79.44} & \textbf{79.28} & \textbf{43.46/60.46} & \textbf{42.72/58.33} & \textbf{32.91/64.84} \\
ZEN 2.0 (L) & \textbf{75.17} & \textbf{82.89} & \textbf{83.09} & 46.99/63.91 & \textbf{46.09/62.37} & \textbf{38.32/70.12} \\
\bottomrule
\end{tabular}
\end{small}
\vspace{-0.3cm}
\end{center}
\caption{\label{result-ar-table}
The overall performance of ZEN 2.0 base (B) and large (L) for Arabic on six NLP tasks with the comparison against our runs of existing pre-trained models (i.e., multilingual BERT, AraBERT, and Arabic BERT).
%(with both base and large version).
%
%The overall performance of Arabic ZEN 2.0 and our runs of different baseline models (i.e., multilingual BERT, AraBERT, and Arabic BERT) on 6 Arabic NLP tasks.
%``(B)'' and ``(L)'' refer to the base and large version.
% We download the previous Arabic models and fine-tune them on all Arabic datasets.
}
\vskip -1em
\end{table*}

In this work, we apply our enhancement of ZEN to two languages, namely, Chinese and Arabic, using much more data than its original version \cite{Sinovation2019ZEN} to ensure its generalization ability.

For Chinese, we follow the common practice in previous studies \cite{devlin2019bert,cui2019pre} to use Chinese Wikipedia dump\footnote{\url{https://dumps.wikimedia.org/zhwiki/}} as one of the corpora.
To expand the generalization ability of ZEN, we use additional large-scale raw text from online resources, including (1) Chinese News Corpus, (2) Chinese Baike Corpus, (3) Chinese Webtext Corpus, (4) Chinese-English Parallel Corpus\footnote{We only use the Chinese part in the corpus.}, (5) Chinese Comments Corpus, and (6) Zhihu Corpus, 
%
% 除了知乎，其他语料主要是从GitHub CLUECorpus2020提供的下载链接获取，CLUEbenchmark这个项目收集了互联网可获取的中文数据，包括中文预训练语料
where the corpora (1)-(5) are obtained from CLUE\footnote{\url{https://github.com/CLUEbenchmark/CLUECorpus2020}} and the corpus (6) is extracted from a well-known Chinese on-line question answering forum\footnote{\url{https://www.zhihu.com}}.
%
% 在数据处理方面，有参考xu2020cluecorpus2020这篇paper，但我们做得更多，需要根据语料做一些定制化的过滤策略
We follow \citet{xu2020cluecorpus2020} to clean the data and perform more operations to improve their quality, such as
%For example, we filter out
filtering out sentences that contain bad words\footnote{\url{https://github.com/LDNOOBW/List-of-Dirty-Naughty-Obscene-and-Otherwise-Bad-Words/}} and non-text content (e.g., HTML mark-ups).
%

%\textcolor{blue}{
%Overall, all corpora contain 268 million sentences and 8 billion Chinese characters in total, where the detailed statistics (the number of sentences and the number of characters) of each corpus is reported in Table \ref{tab:corpus_stat}.}

% 
For Arabic,
% 
% OSCAR官网提到，如果使用他们的语料，请考虑cite他们的paper: suarez2019asynchronous
we collect Arabic News Corpus by crawling multiple Arabic online news websites and download existing resources including Arabic Wikipedia dump\footnote{\url{https://dumps.wikimedia.org/backup-index.html}}, AraCorpus\footnote{\url{http://aracorpus.e3rab.com}}, Tashkeela\footnote{\url{https://sourceforge.net/projects/tashkeela/}} \cite{zerrouki2017tashkeela}, UN Parallel Corpus\footnote{\url{https://conferences.unite.un.org/uncorpus/en/downloadoverview}}, Abu EI-Khair Corpus\footnote{\url{http://www.abuelkhair.net/index.php/en/arabic/abu-el-khair-corpus}} \cite{el20161}, and OSCAR\footnote{\url{https://oscar-corpus.com/}} \cite{suarez2019asynchronous}.
% 
% \textcolor{orange}{The cleaned corpora contains 225M sentences and B tokens in total, which is reported in Table \ref{tab:corpus_stat}.}
% 
% 发现阿语只是简单处理成bert文档格式，没有像中文一样需要多种处理策略
% For all corpora, we \textcolor{red}{how we clean the corpora}.
%
We use BERT tokenizer to segment Arabic text into word-pieces and empirically regard the results as Arabic characters to facilitate character-based encoding in ZEN.

Overall, for the corpora in two languages, the statistics of them are reported in Table \ref{tab:corpus_stat}, with numbers of sentences and tokens presented.
%Table \ref{tab:corpus_stat} reports the number of sentences and the number of characters of each corpus used to train Arabic ZEN 2.0, as well as the total number of them in all corpora.

\subsection{Training}

% architecture
Similar to conventional studies \cite{devlin2019bert,liu2019roberta,wei2019nezha,yang-2019-xlnet,sun2020ernie,baly2020arabert},
we train two versions of the updated ZEN for each language, namely ZEN-base and ZEN-large.
% use Transformer-based encoder as character encoder with a multi-layer n-gram encoder. 
%
For ZEN-base, we use 12 layers of 12-head-self-attention with 768-dimensional hidden vectors for character encoder and 6 layers of 12-head-self-attention for n-gram encoder;
For ZEN-large, we use 24 layers of 16-head-self-attention with 1024-dimensional hidden vectors for character encoder and 6 layers of 16-head-self-attention for n-gram encoder.

Following \citet{Sinovation2019ZEN}, we use pointwise mutual information (PMI) to extract n-grams whose length is in the range $[2, 8]$ from large raw text and filter out rare n-grams according to a frequency threshold to create the n-gram lexicon.
For Chinese and Arabic,
the PMI thresholds are set to 3 and 10, while the n-grams frequency thresholds are set to 15 and 20, respectively.
As a result,
there are 261K and 194K n-grams extracted for Chinese and Arabic, respectively.
For whole n-gram masking in the character encoder, we use WMSeg\footnote{\url{https://github.com/SVAIGBA/WMSeg}.} \cite{tian2020improving} as the off-the-shelf tokenizer to firstly split the input text into n-grams and then combine the adjacent ones into larger n-grams.
%if they appear in the n-gram lexicon.

% 
For both Chinese and Arabic, we train different ZEN models on the obtained large-scale raw text and follow previous studies \cite{devlin2019bert,Sinovation2019ZEN,safaya-etal-2020-kuisail} to optimize them by two semi-supervised tasks, namely, masked language model (MLM) and next sentence prediction (NSP).
% 
% 
% ZEN 2.0是按照从零开始训练的故事说的，lr对标bert应该是1e-4级别。
% 这个地方建议不要说lr=3e-5等信息，容易被人质疑lr为什么是设置这么小。
Following BERT, we use Adam optimizer with warmed-up during the first 36,000 steps and use the learning rate with a peak value of 1e-4 and linear decay.
% 
% In detail, we train ZEN 2.0 with Adam. 
% 
% The learning rate is warmed up over the first 36,000 steps to a peak value of 3e-5 and then linearly decayed. 
% 
The batch size for ZEN-base is set to 24,576, and that for ZEN-large is 8,192.
The total steps of training for Chinese and Arabic are 600K and 800K, respectively.
%we train Chinese ZEN 2.0 and Arabic ZEN 2.0 for 600,000 and 800,000 steps, respectively.
% 
% Chinese and Arabic ZEN 2.0 are trained for 600,000 and 800,000 updates, respectively.
% 
% We use 8K and 24K as batch size for large and base models, respectively.
% 
% 
% the segmentor used for whole n-gram masking 
% 
% number of steps, learning rate, batch size, etc.
% ...
% ...
% We adopt PyTorch with Slurm for distributed training, using 1024 NVIDIA Telsa V100 GPUs with 32GB memory.
% 
% In addition, to speed up the training phase, we adopt mixed precision training using FP16 \cite{micikevicius2017mixed}. 

\section{Fine-tune on Benchmark Tasks}

\subsection{Benchmark Tasks}

To evaluate ZEN 2.0, we fine-tune the models on the benchmark datasets of several different tasks.

For Chinese, we use
Chinese word segmentation (CWS), Part-of-speech (POS) tagging, Named entity recognition (NER),
% \footnote{
% \textcolor{blue}{There are two versions of datasets for NER in the third international Chinese language processing Bakeoff \cite{levow2006third}. ZEN 1.0 uses MSRA and ZEN 2.0 follows previous studies \cite{sun2020ernie,liu2020k} to use MSRA-NER.}},
Document classification (DC), Sentiment analysis (SA), Sentence pair matching (SPM)*, Natural language inference (NLI), Machine reading comprehension (MRC)*, and Question Answering (QA)*, where many of them are introduced in \citet{Sinovation2019ZEN} and we use the same datasets and settings in this work.
Three (marked by *) tasks (datasets) are newly added for ZEN 2.0, with details illustrated below.
% \textcolor{red}{here add items}.
% 
\vspace{-0.15em}
\begin{itemize}[leftmargin=10pt]
    \itemsep-0.15em
    % \item
    % \textbf{NER}:
    % MSRA-NER from the third international Chinese language processing Bakeoff \cite{levow2006third} is used for this task.\footnote{\textcolor{blue}{}}
    \item
    \textbf{SPM}:
    LCQMC \cite{liu2018lcqmc} and the BQ Corpus \cite{chen2018bq} are used.
    \item 
    \textbf{MRC}:
    CMRC 2018 \cite{cui2019span} is used in this task and 
    we evaluate our model performance on its development set following conventional studies \cite{wei2019nezha,sun2020ernie}.
    %of CMRC 2018 \cite{cui2019span}, since this dataset have no official test set.
    %
    \item 
    \textbf{QA}:
     We use the NLPCC-DBQA dataset\footnote{\url{http://tcci.ccf.org.cn/conference/2016/dldoc/evagline2.pdf}} from NLPCC-ICCPOL 2016 Shared Task.
    %  \textcolor{orange}{obtained from NLPCC-ICCPOL 2016}
\end{itemize}
%
%\textcolor{blue}{
%For tasks run in ZEN 1.0, we use the same datasets; for new tasks for Chinese and Arabic, we illustrate the datasets for each task in the following text:
% 
% The following illustrates the additional datasets used in all tasks (for the other datasets for different tasks, we use the same of them as ZEN 1.0):
%}
\vspace{-0.1cm}
For Arabic, we use the following tasks (datasets).
\vspace{-0.1cm}
\begin{itemize}[leftmargin=10pt]
    \itemsep-0.15em
    % 
    % \item 
    % \textbf{Chinese word segmentation (CWS)}*: 
    % CWS is a sequence labeling task only for Chinese ZEN 2.0, where MSR-CWS dataset from SIGHAN2005 Chinese word segmention Backoff \cite{emerson2005second} is used. 
    % 
    \item
    % \textbf{Part-of-speech tagging (POS)}:
    % For Chinese, we use CTB5 \cite{xue2005penn};
    \textbf{POS}: Part 1, 2, and 3 of the Penn Arabic Treebank (ATB)\footnote{
    % We obtain CTB5 dataset from \url{https://catalog.ldc.upenn.edu/LDC2005T01}.
    ATB part 1 is from \url{https://catalog.ldc.upenn.edu/LDC2003T06}, part 2 from \url{https://catalog.ldc.upenn.edu/LDC2004T02}, and part 3 from \url{https://catalog.ldc.upenn.edu/LDC2005T20}.} \cite{maamouri2004penn}.
    \item
    \textbf{NER}:
    % For Chinese, we use MSRA-NER dataset from Microsoft Research Asia \cite{levow2006third}; 
    AQMAR \cite{mohit2012recall} and ANERCorp \cite{benajiba2007anersys} containing articles from Wikipedia and newswire, respectively.
    % AQMAR dataset contrains 28 Arabic Vikipedia articles, and ANERCorp dataset is selected from news wire and other type web sources.
    % with xx and xx unique named entity types, respectively.
    % 
    \item
    \textbf{DC}:
    AR-5, AB-7, and KH-7 from the SANAD \cite{einea2019sanad} dataset, containing 5, 7, and 7 unique document types, respectively.
    % \textcolor{orange}{, and collected from three Arabic news portals which are AlKhaleej, AlArabiya, and Akhbarona, respectively.}
    % dataset, which contain 5, 7, and 7 types of categories, respectively.
    % For Chinese, we use THUCNews dataset \cite{sun2016thuctc} from Sina News containing 10 types of documents; 
    % We use Arabic SANAD \cite{einea2019sanad} dataset which consists of three sub-datasets, i.e.,
    % containing 5, 7, and 7 types of categories, and collected from three Arabic news portals which are AlKhaleej, AlArabiya, and Akhbarona, respectively.
    % \textcolor{red}{why mention the three subsets?}
    % 
    \item
    \textbf{SA}:
    The ASTD \cite{nabil2015astd} dataset that contains around 10,000 tweets.
    % 
    % \item
    % \textbf{Sentence pair matching (SPM)}*:
    % SPM is a sentence pair classification task only for Chinese ZEN 2.0, where LCQMC \cite{liu2018lcqmc} and BQ Corpus \cite{chen2018bq} are used.
    % 
    \item
    \textbf{NLI}:
    % NLI is another sentence pair classification task. 
    % For Chinese, we use the Chinese part of the XNLI \cite{conneau2018xnli}; 
    The Arabic part of the XNLI \cite{conneau2018xnli} is used for this task.
    \item
    \textbf{MRC}:
    %  MRC aims to extract an answer span in a given document for the corresponding question.
    %  We use 
    %  CMRC 2018 \cite{cui2019span} for Chinese and 
     Arabic-SQuAD \cite{mozannar2019neural} and ARCD \cite{mozannar2019neural}.
    %  for Arabic.
    %  Note that CMRC 2018 does not have an official test set; evaluation is performed on the official development set in previous studies.
     % 
\end{itemize}

For all datasets used in the experiments, we follow previous studies to pre-process them and split them into train/dev/test sets.
% 
% The detailed data split information and statistics (i.e., the number of sentences and tokens) are introduced in Appendix \textcolor{red}{xx}.
% 
We follow the common practice to evaluate the performance of all models, i.e.,
we use F1 scores
for CWS and NER,
and use accuracy
for POS tagging, DC, SA, SPM, and NLO;
we use both exact match (EM) and F1 scores for MRC;
and for QA, we use mean reciprocal rank (MRR) and F1 scores.
For each dataset of a particular task, we fine-tune ZEN 2.0 on the training set and evaluate it on the test set\footnote{We follow previous studies \cite{wei2019nezha,sun2020ernie} to evaluate Chinese ZEN 2.0 on the development set of CMRC2018 since it does not have an official test set.}.

% \begin{figure*}[t]
% \begin{subfigure}{.5\textwidth}
%   \centering
%   \includegraphics[width=0.98\textwidth, trim=0 20 0 0 ]{figures/pretrain_step_xnli.png}
%   \caption{\textcolor{orange}{}}
%   \vskip -1em
% \end{subfigure}
% \begin{subfigure}{.5\textwidth}
%   \centering
%   \includegraphics[width=0.98\textwidth, trim=0 20 0 0 ]{figures/pretrain_step_cmrc2018.png}
%   \caption{\textcolor{orange}{}}
%   \vskip -1em
% \end{subfigure}
% \label{fig: pretrain steps}
% \caption{\textcolor{orange}{Performance of XNLI (a) and CMRC2018 (b) fine-tuned on different large BERT, ZEN 1.0 and ZEN 2.0 checkpoint models pre-training at different steps.}}
% \end{figure*}

\begin{figure*}[t]
  \centering
  \includegraphics[width=0.98\textwidth, trim=0 20 0 0 ]{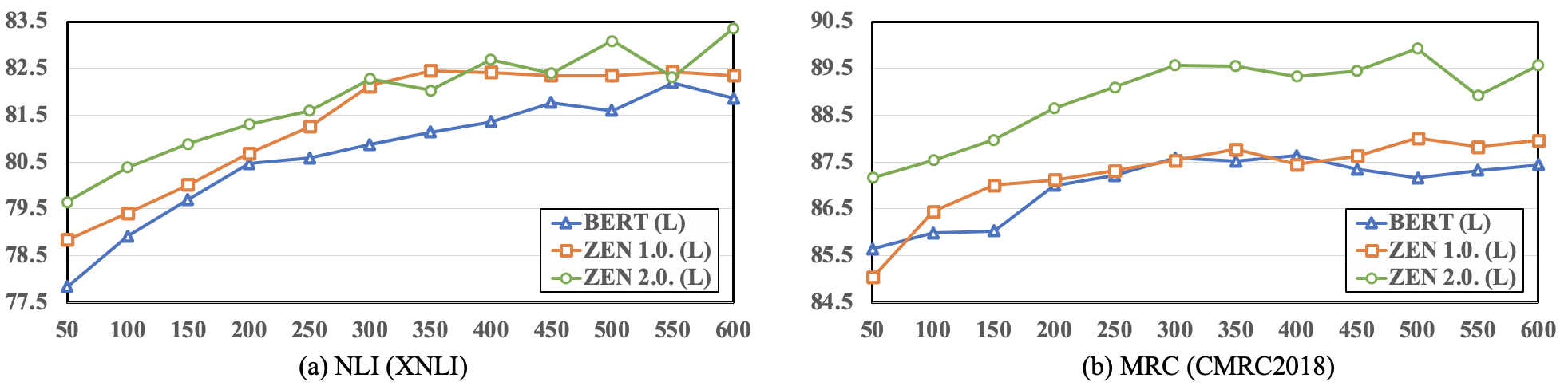}
  \caption{
  The performance of different models on NLI (a) and MRC (b) with respect to the number of pre-training steps (in thousands), where
  the curves of BERT (L), ZEN 1.0 (L), and ZEN 2.0 (L) are illustrated in blue, orange, and green colors, respectively.
  The evaluation metric for NLI is accuracy and that for MRC is the F1 score.}
  \vskip -1em
  \label{fig: pretrain steps}
\end{figure*}

% \begin{figure*}[t]
%   \centering
%   \includegraphics[width=0.98\textwidth, trim=0 20 0 0 ]{figures/pretrain-n-gram.png}
%   \caption{\textcolor{blue}{}}
%   \vskip -1em
%   \label{fig: ngram tsne}
% \end{figure*}

\subsection{Overall Results}

Table \ref{result-zh-table} reports the performance of base (B) and large (L) Chinese ZEN 2.0 on different tasks and the comparison with previous representative Chinese text encoders, 
including ERNIE 1.0 and 2.0 \cite{sun2019ernie,sun2020ernie}, RoBERTa-WWM (i.e., RoBERTa with whole word masking) \cite{cui2019pre}, NEZHA-WWM (i.e., NEZHA with whole word masking) \cite{wei2019nezha}, MWA \cite{li-etal-2020-enhancing}, 
% ERNIE 2.0 \cite{sun2020ernie}, 
MacBERT \cite{cui2020revisiting}
as well as ZEN 1.0, where ZEN 2.0 achieves the highest performance on all tasks.
Table \ref{result-ar-table} reports the performance of Arabic ZEN 2.0 (base and large version) compared with other widely used Arabic text encoders (some are from our runs), namely, multilingual BERT \cite{devlin2019bert}, AraBERT 0.1 \cite{baly2020arabert}, and Arabic BERT \cite{safaya-etal-2020-kuisail},
where both versions of Arabic ZEN 2.0 outperform their corresponding baselines on all tasks.

A general summary from the results can be drawn that, 
% \textcolor{blue}{Overall},
n-gram information works well with ZEN 2.0 in different sizes (i.e., the base and large version),
where refined n-gram representation, whole n-gram masking, and character encoding with relative positional information well collaborated with each other and improve the performance of ZEN 2.0 on different NLP tasks.
Specifically, although directly upgrading ZEN 1.0 from base to large version improves its performance on many NLP tasks, ZEN 2.0 can be further boosted, demonstrating the necessity of the proposed enhancements.
In addition, compared with previous pre-trained models that learn word (n-gram) information through different masking strategies, ZEN 2.0 is able to explicitly encode n-gram information in a more effective manner through both the refined n-gram representation and the whole n-gram masking, which leads ZEN 2.0 to outperform all previous studies as well as ZEN 1.0 on all tasks.
%
% In addition,
Moreover,
even though the languages are highly different between Chinese and Arabic,
n-gram information is proved to be helpful for Arabic as well, although ZEN is not originally designed for it.

\begin{figure}[t]
  \centering
  \includegraphics[width=0.48\textwidth, trim=0 20 0 0 ]{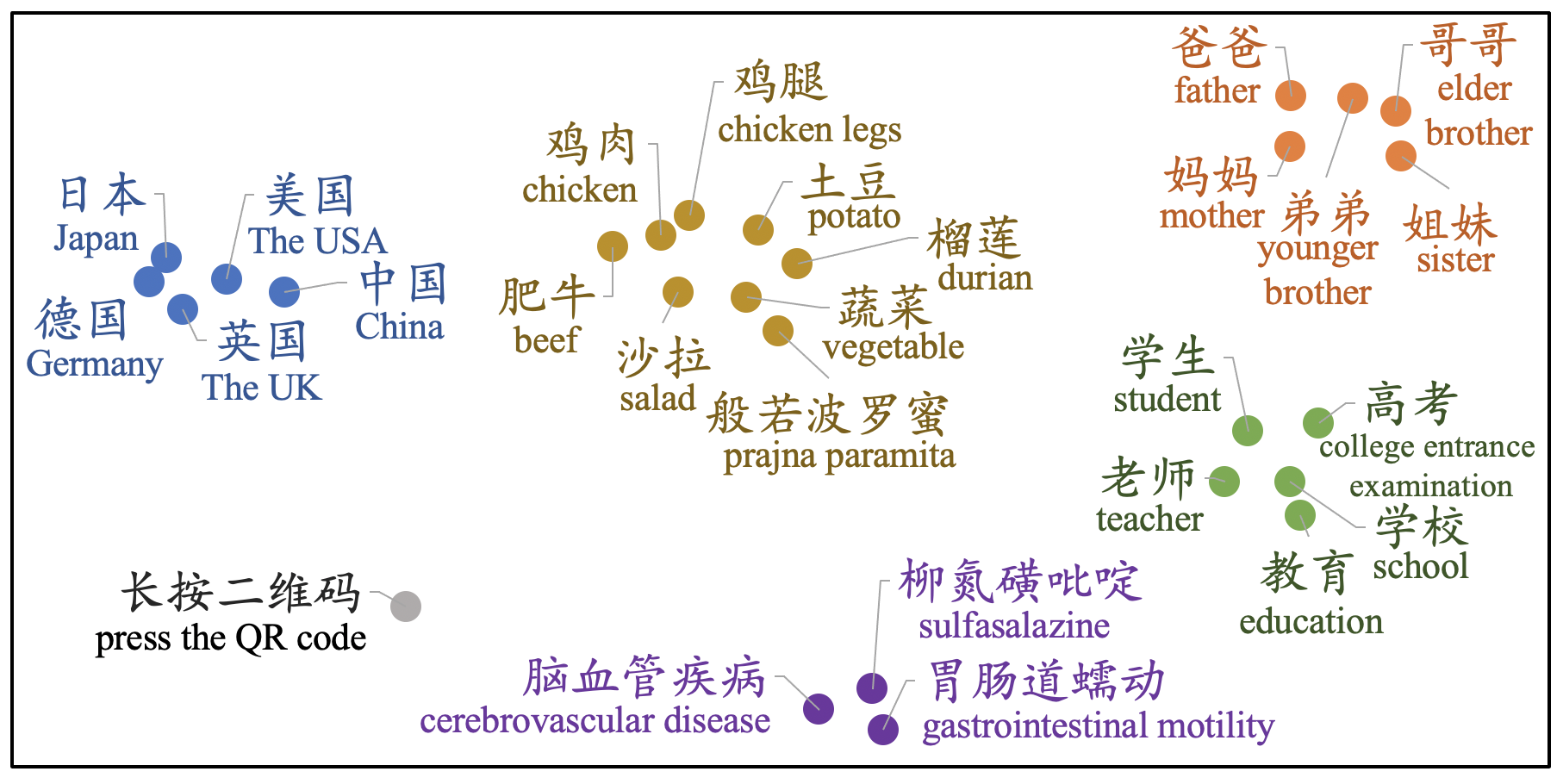}
  \caption{
  Visualization of n-gram representations for some examples.
  The distance between two n-grams illustrates the similarity between their representations, where a low distance indicates the two n-grams have similar representations.
  N-grams in the same cluster are represented in the same color.
%   Two-dimensional T-SNE projection of the 512-dimensional N-gram representation. N-grams of same category are visualized in blue color, N-grams that have relations are visualized in green color.
  }
  \vskip -1em
  \label{fig: ngram tsne}
\end{figure}

\section{Analysis}

% 有跨语言的分析结果
% 证明组合ngram的方式对不同语言也适用
% 同时当模型尺寸变大的时候ngram表征也同样有效

% \input{tables/analysis_ngram}

\begin{table*}[h]
\begin{center}
\begin{small}
\begin{tabular}{p{1.8cm}
                   L{1.1cm}
                   C{1.7cm} 
                   C{2.0cm}  
                   C{1.7cm} 
                   C{1.7cm} 
                   C{2.3cm} }
\toprule
\multirow{3}*{\textbf{}}
&\multirow{3}*{\textbf{WNM}}
&\multicolumn{1}{c}{\textsc{\textbf {CWS}}}
&\multicolumn{1}{c}{\textsc{\textbf{POS tagging}}}
&\multicolumn{1}{c}{\textsc{\textbf{NER}}}
&\multicolumn{1}{c}{\textsc{\textbf{DC}}}
&\multicolumn{1}{c}{\textsc{\textbf{SA}}}\\
\addlinespace[0.05cm]
\cline{3-7}
\addlinespace[0.05cm]
~& ~
&\multicolumn{1}{c}{\textsc{\textbf {MSR-CWS}}}
&\multicolumn{1}{c}{\textsc{\textbf{CTB5}}}
&\multicolumn{1}{c}{\textsc{\textbf{MSRA-NER}}}
&\multicolumn{1}{c}{\textsc{\textbf{THUCNews}}}
&\multicolumn{1}{c}{\textsc{\textbf{ChnSentiCorp}}}\\
\addlinespace[0.05cm]
\cline{3-7}
\addlinespace[0.05cm]
~& ~& \multicolumn{1}{c}{\small{F1}} &
\multicolumn{1}{c}{\small{ACC}} &
\multicolumn{1}{c}{\small{F1}} &
\multicolumn{1}{c}{\small{ACC}} &
\multicolumn{1}{c}{\small{ACC}} \\
\midrule	
% ZEN (B) & N/A & 98.36 & 96.91 & \textbf{95.54} & 97.72 & \textbf{96.67} \\
%  & Jieba & 98.40 & 96.97 & 95.44 & \textbf{97.76} & 96.17 \\
%  & WMSeg & \textbf{98.41} & \textbf{97.00} & \textbf{95.54} & 97.64 & 96.08 \\
% \midrule
\multirow{3}*{ZEN 2.0 (L)} & WMSeg & \textbf{98.66} & \textbf{97.09} & \textbf{96.20} & \textbf{97.93} & \textbf{96.50} \\
& Jieba & 98.57 & 96.99 & 96.18 & 97.90 & 96.33 \\
& N/A & 98.52 & 96.92 & 96.08 & 97.90 & 96.17 \\
\bottomrule
\end{tabular}
\begin{tabular}{p{1.8cm}
                   L{1.1cm}
                   C{1.8cm} 
                   C{1.8cm} 
                   C{1.85cm} 
                   C{2.1cm} 
                   C{2.35cm} }
\toprule
\multirow{3}*{\textbf{}}
&\multirow{3}*{\textbf{WNM}}
&\multicolumn{2}{c}{\textsc{\textbf{SPM}}}
&\multicolumn{1}{c}{\textsc{\textbf{NLI}}}
&\multicolumn{1}{c}{\textsc{\textbf{MRC}}}
&\multicolumn{1}{c}{\textsc{\textbf{QA}}}\\
\addlinespace[0.05cm]
\cline{3-7}
\addlinespace[0.05cm]
~ & ~
&\multicolumn{1}{c}{\textsc{\textbf{LCQMC}}}
&\multicolumn{1}{c}{\textsc{\textbf{BQ Corpus}}}
&\multicolumn{1}{c}{\textsc{\textbf{XNLI}}}
&\multicolumn{1}{c}{\textsc{\textbf{CMRC2018}}}
&\multicolumn{1}{c}{\textsc{\textbf{NLPCC-DBQA}}}\\
\addlinespace[0.05cm]
\cline{3-7}
\addlinespace[0.05cm]
~& ~& \multicolumn{1}{c}{\small{ACC}} &
\multicolumn{1}{c}{\small{ACC}} &
\multicolumn{1}{c}{\small{ACC}} &
\multicolumn{1}{c}{\small{EM/F1}} &
\multicolumn{1}{c}{\small{MRR/F1}} \\
\midrule	
% ZEN (B) & N/A & 88.20 & 85.23 & 77.78 & 66.70/85.89 & 94.31/83.08 \\
%  & Jieba & 88.18 & 85.31 & 80.20 & 69.71/87.52 & 95.45/84.07 \\
%  & WMSeg & \textbf{88.39} & \textbf{85.65} & \textbf{80.28} & \textbf{70.77/87.97} & \textbf{95.74/84.43} \\
% \midrule
\multirow{3}*{ZEN 2.0 (L)} & WMSeg & \textbf{88.81} & \textbf{85.99} & \textbf{83.09} & \textbf{73.00/89.92} & \textbf{96.11/86.47} \\
& Jieba & 88.54 & 85.67 & 82.44 & 71.76/88.95 & 95.71/86.19 \\
& N/A & 88.48 & 85.70 & 81.06 & 70.58/87.84 & 95.58/86.35 \\
\bottomrule
\end{tabular}
\end{small}
\vspace{-0.3cm}
\end{center}
\caption{\label{result-seg-zh-table}
The performance of ZEN 2.0 large for Chinese with whole n-gram masking (WNM) when different off-the-shelf tokenizers (i.e., WMSeg and Jieba) are used. ``N/A'' standards for that the model with character masking.
% The performance of ZEN without and with whole word masking using different segmentor on 9 Chinese NLP tasks. Only CMRC2018 dataset reports the result of development dataset, the other datasets report the results of test dataset.
}
\vskip -0.5em
\end{table*}

\subsection{The Effect of Training Steps}

% 不同step模型ppl、xnli、cmrc2018、msra-ner、bq corpus 任务ft的performance
% 说明：不是某个step的checkpoint在所有任务上的表现都是最好的
To analyze the effect of the enhancement on ZEN,
we use two different tasks (i.e., NLI and MRC) to demonstrate the
performance of ZEN 2.0 during the pre-training process.
% The number of training steps is one of the most important factors that influence the performance of ZEN 2.0.
%
% In this analysis, 
Specifically, we use ZEN 2.0 and the baseline models (i.e., ZEN 1.0 and BERT) for Chinese at different pre-training steps and then fine-tune them on the XNLI and CMRC2018 datasets.
The curves of the performance (i.e., accuracy for NLI and F1 scores for MRC) on the two tasks with respect to the pre-training steps (in thousands) are illustrated in Figure \ref{fig: pretrain steps} (a) and Figure \ref{fig: pretrain steps} (b), respectively.
It can be observed from the results that, for both tasks, ZEN 2.0 outperforms the two baselines at different pre-training steps, particularly when the training is at the early stage (e.g., when the number of training steps in fewer than 100K).
This observation confirms that the enhancements proposed in this work help the training of ZEN, when model size is enlarged, ZEN 2.0 is able to generate a good text representation in a more effective way, especially for tasks like NLI and MRC that normally require high-level understanding of the input texts.

\subsection{The Effect of N-gram Representations}
%同样语料自己训练的BERT+ZEN 2.0，base+large -> 当模型尺寸变大的时候ngram表征也同样有效
% 中文+阿语 -> ngram的方式对不同语言也适用

N-grams are very useful features to represent contextual information and they are explicitly encoded in ZEN.
%by the n-gram encoder in ZEN 2.0.
%
We already show that in Table \ref{result-zh-table} and Table \ref{result-ar-table}, with n-gram representations, the pre-trained models (both ZEN 1.0 and 2.0) outperform other ones without such mechanism on different NLP tasks.
Therefore, it is interesting to analyze the n-gram representations by qualitatively investigating their relations, which is similar to that has been done for word embeddings.
%Therefore, similar to word embeddings, the quality of n-gram representation highly influences the quality of text representation generated by ZEN 2.0, which can further significantly influence the model performance on different NLP tasks.
% 
%To explore the effect of n-gram representations,
In doing so,
% we firstly run ZEN 2.0 large on the \textcolor{blue}{pre-training} dataset and then 
we collect the n-gram representations from the first layer of the n-gram encoder in ZEN 2.0.
Then, for each n-gram, we average its representation vectors under different contexts and regard the resulting vector as the final n-gram representations.
Figure \ref{fig: ngram tsne} visualizes the final representations of some example n-grams, where the distance between two n-grams indicates their similarity (lower distances indicate the n-grams are more relevant).
It is observed that n-grams with relevant semantic meanings are grouped into the same cluster (n-grams in different clusters are represented in different colors), while the irrelevant ones are far away from each other.
For example, ``\begin{CJK}{UTF8}{gkai}美国\end{CJK}'' (\textit{the USA}), ``\begin{CJK}{UTF8}{gkai}中国\end{CJK}'' (\textit{China}),
``\begin{CJK}{UTF8}{gkai}英国\end{CJK}'' (\textit{the UK}), ``\begin{CJK}{UTF8}{gkai}德国\end{CJK}'' (\textit{Germany}), and ``\begin{CJK}{UTF8}{gkai}日本\end{CJK}'' (\textit{Japan}) that all represent countries, are in the same cluster (represented in blue color), while they are far away from irrelevant n-grams, e.g., 
``\begin{CJK}{UTF8}{gkai}柳氮磺吡啶\end{CJK}'' (\textit{sulfasalazine}).
% ``\begin{CJK}{UTF8}{gkai}般若菠萝蜜\end{CJK}'' (\textit{prajna paramita}).}
%
%This observation demonstrates the effectiveness of ZEN 2.0 to learn n-grams representations.
This finding is inspiring since the n-grams are automatically generated so that the learning process for ZEN shows its validity in assigning their representations with proper values and ensuring that their relevance in semantics are appropriately modeled.
% which results in the similar n-gram representations.
Training ZEN makes it possible to learn embeddings for larger granular text (e.g., phrases) without explicit extraction of them.

\subsection{The Effect of Whole N-gram Masking}

% 中文
% 分析wwm的作用
% 对比不同segmentor的作用
Whole word masking is proved to be useful in learning many previous pre-trained models.
However, words are hard to be identified in Chinese (and Arabic in many cases),
we therefore use n-gram masking instead in this work.
Since we use a word segmenter to tokenize input text into pieces as the first step and then combine some pieces into larger n-grams for masking,
the performance of the segmenter is vital for obtaining reasonable n-grams.
To illustrate the effectiveness of using WMSeg as the segmenter, in this analysis,
%
%For Chinese ZEN 2.0,
we compare ZEN 2.0 (large version)
%trained with and without whole n-gram masking (WNM) and test on all Chinese tasks with
trained with whole n-gram masking (WNM) when a different segmenter, i.e., Jieba\footnote{ \url{https://github.com/fxsjy/jieba}. We choose Jieba as the comparing segmenter for its widely usage as the conventional tool for 
% word masking in previous Chinese pre-trained models \cite{wei2019nezha} \textcolor{red}{(citations)}.
Chinese word segmentation in many previous studies \cite{wei2019nezha,chen-etal-2020-neural}.}, is applied with the same n-gram lexicon.
% fine-tune the models on different tasks.
% 
% In addition, to explore the effect of different word segmentor, in the training process, we 
%using different segmenters, namely, Jeiba\footnote{We use Jieba from \url{https://github.com/fxsjy/jieba}.} and WMSeg \cite{tian2020improving}, with the same n-gram lexicon 
% for the whole n-gram masking, 
%to obtain the segmented n-grams in the input text.
% 
% conduct several experiments of Chinese ZEN models without and with whole word masking using Jieba\footnote{We use Jieba from \url{https://github.com/fxsjy/jieba}.} and WMSeg \cite{tian2020improving} as segmentor, respectively.
% 
Table \ref{result-seg-zh-table} reports the comparison (ZEN 2.0 large) on all Chinese tasks.
% 
% For all the tasks, only CMRC2018 dataset reports the result of development dataset, the other datasets report the results of test dataset.
%
There are several observations.
First, for all tasks, ZEN 2.0 with WNM obtains higher results than the models with character masking, which
complies with the findings in previous studies \cite{sun2019ernie,cui2019pre,wei2019nezha,sun2020ernie,cui2020revisiting}.
%confirms the effectiveness of WNM to improve Chinese ZEN 2.0.
% can help improve ZEN 2.0 for all tasks, where most of the downstream tasks fine-tuned on ZEN models using WMSeg as segmentor for whole word masking achieve higher scores than the ones without whole word masking, especially tasks required higher ability of understanding, such as NLI, MRC, and QA.
%
Second, when using WNM, WMSeg outperforms Jieba and this option achieves the highest results on all tasks.
%
% Second, better segmentor for whole word masking can help ZEN models achieve higher performance on downstream tasks, where the results of downstream task fine-tuned on Chinese ZEN models using WMSeg as segmentor for whole word masking achieve higher scores than the ones using Jieba,
% 
% since WMSeg achieve higher performance than Jieba on Chinese word segmention, one possible explanation could be that WMSeg can provide more accurate cures about the boundary information of n-grams and help improve model's ability of encoding contextual information. 
%
This observation indicates that a better segmenter is of great importance for masking larger context,
%which is 
because WMSeg provides 
%The explanation could be that WMSeg has better performance on Chinese tokenization than Jieba, which allows WMSeg to provide 
more accurate information of word boundaries so that the masked n-grams tend to be more reasonable semantic units.
%more useful contextual information.
% information of n-grams and help improve model's ability of encoding contextual information. since WMSeg achieve higher performance than Jieba on Chinese word segmention, 

\begin{figure}[t]
  \centering
  \includegraphics[width=0.48\textwidth, trim=0 20 0 0 ]{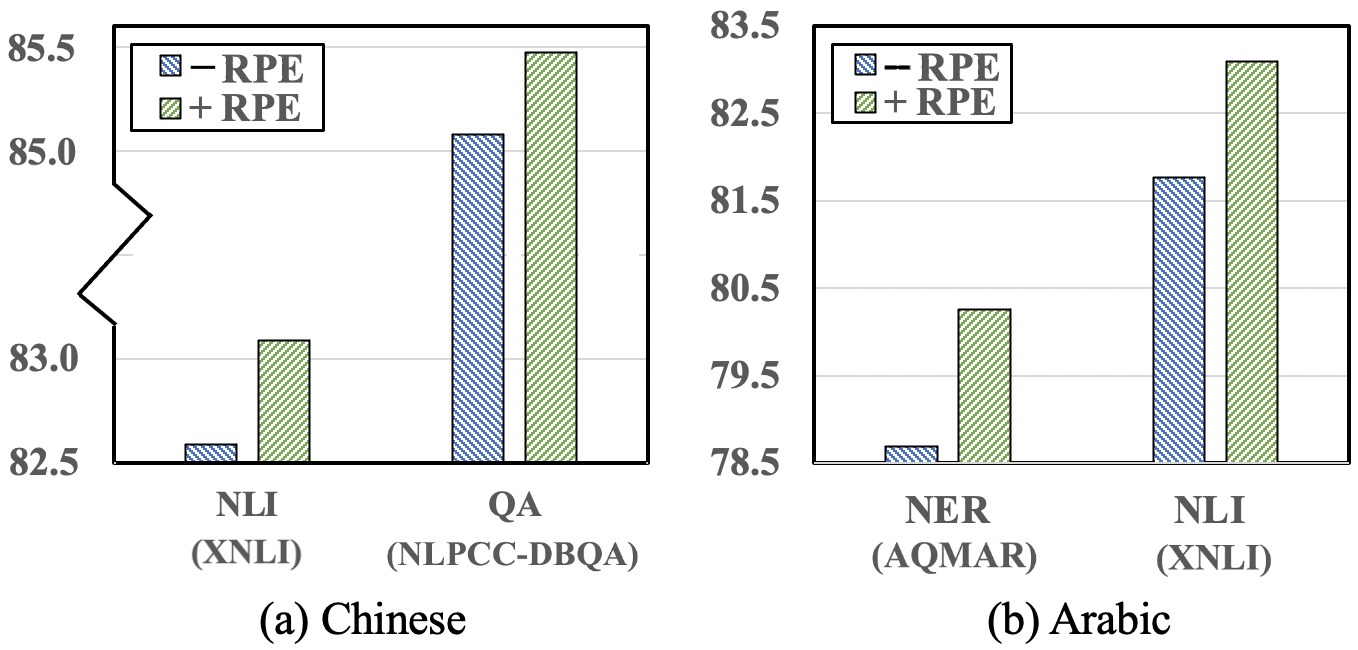}
  \caption{
  The performance histograms of ZEN 2.0 with (+) and without (-) relative positional encoding (RPE) on different Chinese (a) and Arabic (b) NLP tasks.
%   on Chinese (a) NLI and QA, and on Arabic (b) NER and NLI.
%   
  The evaluation metric for NER and QA is F1 scores and that for NLI is the accuracy.
  }
  \vskip -1em
  \label{fig: pretrain adapted}
\end{figure}

\subsection{Relative Positional Encoding Effect}

% 中文
% 分析Adapted的作用
%
To address the limitation of the vanilla Transformer in character encoding (which is used by ZEN 1.0),
%ZEN 1.0 and BERT use vanilla Transformer for character encoding, which fails to capture the relative positional information between input characters.
% 
%To address this limitation,
ZEN 2.0 applies the relative positional encoding technique to the character encoder.
To explore the effect of this enhancement, we compare the performance of ZEN 2.0 with and without relative positional encoding (RPE) on two Chinese NLP tasks (i.e., NLI and QA) and two Arabic NLP tasks (i.e., NER and NLI).
%
% To analysis the effects of using adapted character encoding for ZEN models, we conduct comparison experiments for adapted ZEN models and vanilla ZEN models.
%
Figure \ref{fig: pretrain adapted} shows the comparison between the models, in which the one with RPE (+RPE) consistently outperforms the one without RPE (-RPE) on all chosen tasks, demonstrating the effectiveness of modeling relative position information.
Particularly, more than 2\% improvement on the F1 score is observed on AQMAR dataset for Arabic NER task.
% and can further enhance the text representation. 
% of downstream tasks fine-tuned on Chinese ZEN models.
% 
% For all the tasks, only CMRC2018 dataset reports the result of development dataset, the other datasets report the results of test dataset.
% 
% From the results, it can be seen that the results of downstream tasks fine-tuned on adapted ZEN models achieve higher scores than the ones of vanilla ZEN models, both large and base models,
% 
% which indicates that adapted character encoding can improve ZEN models performance for downstream tasks.
%
It can be explained that the relative position information of Arabic texts is more important (in most cases the morphology of an Arabic word is different according to its position in a sentence);
ZEN 2.0 with RPE is able to capture that information and thus generates high-quality text representations, which can further improve model performance on Arabic NLP tasks.

% 如何抽取数据？
% 首先，提取xnli数据集的句子，用ZEN 2.0 (L)作为encoder，
% 提取句子中match的ngram，
% 用encoding最后一层的输出提取出ngram的hidden state，
% 经过average pooling得到ngram feature。
% 一个ngram可能会有多个不同句子得到的feature，进行mean处理，得到该ngram唯一feature
% 使用skearn的TSNE，输入ngram的feature数据
% 故事：
% 1、同类别的词聚在一起，譬如国家名字 ，
% 2、有关系的词聚在一起，譬如大学相关的词，包括教授、学院等

\subsection{Case Study}

% 这里挑选XNLI任务来做case，分析ZEN 2.0和BERT的对比
% 挑选这么一个case，加了ngram信息之后，模型能够把attention更多放在关键的词上，并且关键的词能够提供帮助模型识别推理关系

\begin{figure}[t]
  \centering
  \includegraphics[width=0.48\textwidth, trim=0 25 0 0 ]{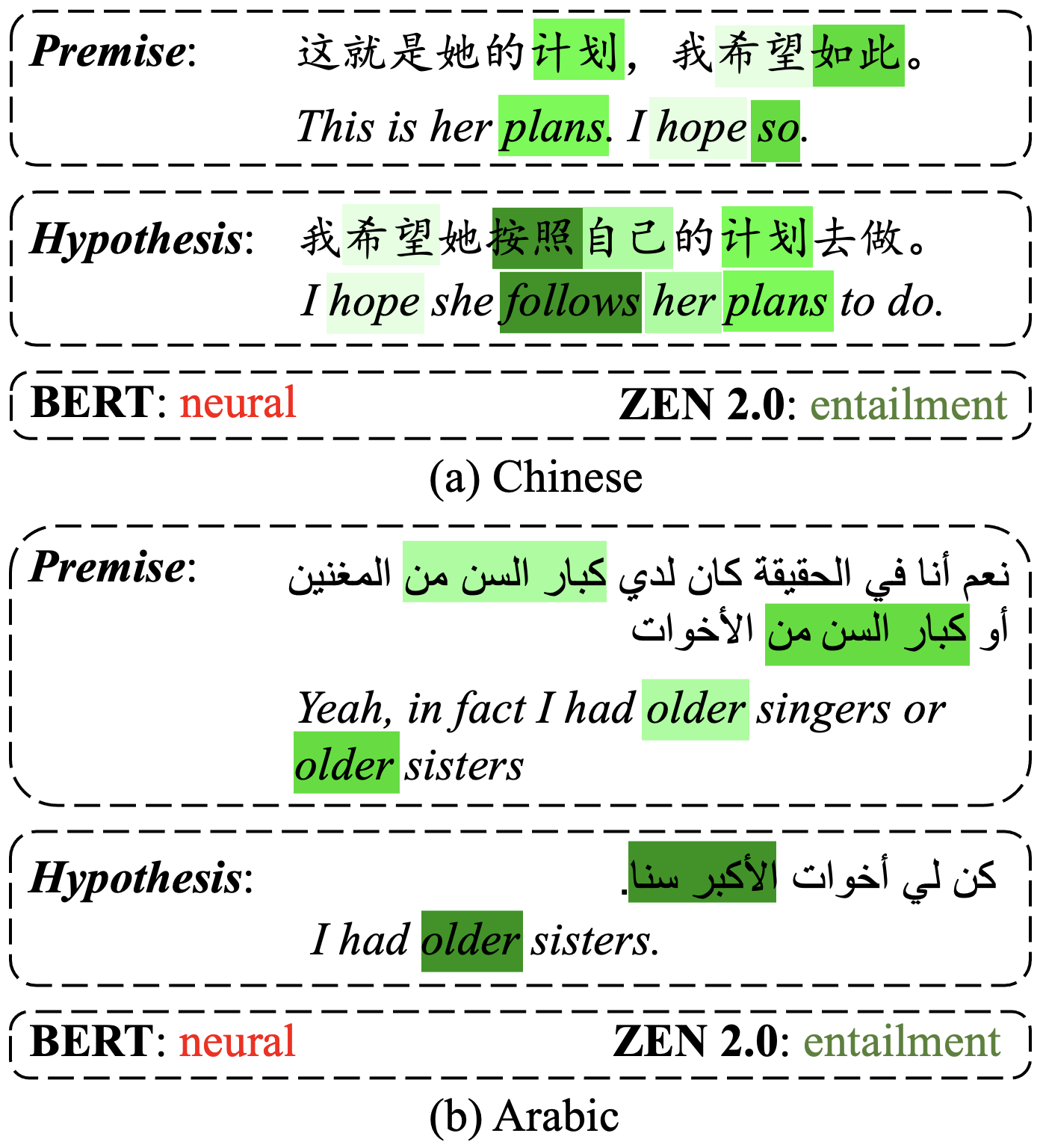}
  \caption{
  A case study on the NLI task with two examples from Chinese (a) and Arabic (b).
  For both examples, ZEN 2.0 correctly predicts that the premise entails the hypothesis, while BERT fails to do so.
  The weight assigned to different n-grams in the n-gram encoder of ZEN 2.0 is visualized by different colors on the corresponding n-grams and their English translations, with deeper colors referring to higher weights.
  }
  \vskip -1em
  \label{fig: case study}
\end{figure}

%
To further examine how ZEN 2.0 leverages n-gram information to improve model performance, we conduct a case study on the NLI task, which is a difficult one requiring models to have a good understanding of contextual information in order to make correct predictions.
Figure \ref{fig: case study} shows two example (\textit{premise}, \textit{hypothesis}) pairs from Chinese and Arabic, where, for both examples, ZEN 2.0 successfully predicts that the text entailment relation between the premise and the hypothesis is ``\textit{entailment}'', while the BERT baseline model fails to do so.
In addition, we visualize the attentions assigned to different n-grams in the n-gram encoder of ZEN 2.0 on their corresponding n-grams (as well as the English translations) in the premise and the hypothesis in different colors, where darker colors refer to higher weights.
%
% \begin{CJK}{UTF8}{gkai}按照\end{CJK}
In the Chinese example (i.e., Figure \ref{fig: case study}) (a), ZEN 2.0 successfully distinguishes the importance of different n-grams and assigns higher weights to ``\begin{CJK}{UTF8}{gkai}按照\end{CJK}'' (``\textit{follow}''), ``\begin{CJK}{UTF8}{gkai}如此\end{CJK}'' (``\textit{so}''), and ``\begin{CJK}{UTF8}{gkai}计划\end{CJK}'' (``\textit{plans}'') that provides strong cues indicating the premise entails the hypothesis.
Similarly, in the Arabic example (i.e., Figure \ref{fig: case study} (a)), ``\RL{kbAr Alsn mn}'' (``\textit{older}'') in the premise and ``\RL{AlOkbr snA}'' (``\textit{older}'') in the hypothesis obtain high weights.
Thus, ZEN 2.0 is able to leverage the information from highlighted n-grams (which are essential in NLI) to make correct predictions, while BERT is unable to do so and predicts the incorrect results.

\section{Conclusion}
We propose ZEN 2.0, an updated n-gram enhanced pre-trained encoder on Chinese and Arabic,
with different improvements such as refined n-gram representations, whole n-gram masking and relative positional encoding applied to ZEN 1.0 and enlarged model size corresponding to BERT-large.
Compared to its previous version, ZEN 2.0 outperforms it on all tested NLP tasks, including several new ones added to the fine-tune list.
Moreover, for both Chinese and Arabic NLP tasks, ZEN 2.0 shows its superiority to other existing representative pre-trained models by achieving the state-of-the-art performance.
Analyses are also conducted to investigate the effect of different improvements, where the findings further demonstrate the effectiveness of them in improving the representation ability of ZEN 2.0.
%\textcolor{red}{some words about case study}
The case study conducted on Chinese and Arabic NLI task confirms that ZEN 2.0 appropriately leverages n-gram information to achieve a good understanding of the input text and thus obtain promising performance on this task.

% \bibliography{anthology,acl2020}
\bibliography{acl2021}
\bibliographystyle{acl_natbib}

\end{document}